\documentclass{article}

\usepackage{microtype}
\usepackage{graphicx}
\usepackage{subcaption}
\usepackage{caption}

\usepackage{booktabs} %

\usepackage{hyperref}

\usepackage[accepted]{icml2024}

\usepackage{amsmath}
\usepackage{amssymb}
\usepackage{mathtools}
\usepackage{amsthm}

\usepackage[capitalize,noabbrev]{cleveref}

\theoremstyle{plain}

\theoremstyle{remark}

\newtheorem{definition}{Definition}
\newtheorem{observation}{Observation}

\usepackage[textsize=tiny]{todonotes}

\def\Snospace~{\S{}}

\def\Snospace~{\S{}}

\usepackage{bm}

\usepackage{algorithm}
\usepackage{algorithmicx}
\usepackage[noend]{algpseudocode}
\usepackage{comment}
\algnewcommand\algorithmicinput{\textbf{Input:}}
\algnewcommand\Input{\item[\algorithmicinput]}
\algnewcommand\algorithmicoutput{\textbf{Output:}}
\algnewcommand\Output{\item[\algorithmicoutput]}
\algnewcommand{\LineComment}[1]{\State \(\triangleright\) #1}

\usepackage{xspace}
\newcommand{\sys}{\mbox{\textsc{LatentTracer}}\xspace}

\usepackage{booktabs}
\usepackage{multirow}
\usepackage{graphicx}
\usepackage{float}
\usepackage[super]{nth}
\usepackage{pifont}
\usepackage{makecell}
\usepackage{longtable}

\icmltitlerunning{How to Trace Latent Generative Model Generated Images without Artificial Watermark?}

\begin{document}

\twocolumn[
\icmltitle{How to Trace Latent Generative Model Generated Images without Artificial Watermark?}

\begin{icmlauthorlist}
\icmlauthor{Zhenting Wang}{Rutgers University}
\icmlauthor{Vikash Sehwag}{Sony AI}
\icmlauthor{Chen Chen}{Sony AI}
\icmlauthor{Lingjuan Lyu}{Sony AI}
\icmlauthor{Dimitris N. Metaxas}{Rutgers University}
\icmlauthor{Shiqing Ma}{University of Massachusetts Amherst}

\end{icmlauthorlist}

\icmlaffiliation{Rutgers University}{Rutgers University}
\icmlaffiliation{Sony AI}{Sony AI}
\icmlaffiliation{University of Massachusetts Amherst}{University of Massachusetts at Amherst}

\icmlcorrespondingauthor{Lingjuan Lyu}{Lingjuan.Lv@sony.com}

\icmlkeywords{Machine Learning, ICML}

\vskip 0.3in
]

\printAffiliationsAndNotice{\icmlIntern}

\begin{abstract}

Latent generative models (e.g., Stable Diffusion) have become more and more popular, but concerns have arisen regarding potential misuse related to images generated by these models. It is, therefore, necessary to analyze the origin of images by inferring if a particular image was generated by a specific latent generative model. Most existing methods (e.g., image watermark and model fingerprinting) require extra steps during training or generation. These requirements restrict their usage on the generated images without such extra operations, and the extra required operations might compromise the quality of the generated images.
In this work, we ask whether it is possible to \emph{effectively and efficiently} trace the images generated by a specific latent generative model without the aforementioned requirements. To study this problem, we design a latent inversion based method called \sys to trace the generated images of the inspected model by checking if the examined images can be well-reconstructed with an inverted latent input. We leverage gradient based latent inversion and identify a encoder-based initialization critical to the success of our approach. Our experiments on the state-of-the-art latent generative models, such as Stable Diffusion, show that our method can distinguish the images generated by the inspected model and other images with a high accuracy and efficiency. Our findings suggest the intriguing possibility that today's latent generative generated images are naturally watermarked by the decoder used in the source models. Code: \url{https://github.com/ZhentingWang/LatentTracer}.

\end{abstract}
    
\section{Introduction}
\label{sec:intro}

Recently, latent generative models~\citep{rombach2022high} have 
attracted significant attention and
showcased outstanding capabilities in generating a wide range of high-resolution images with surprising quality.
Many state-of-the-art image generation models belong to \textit{latent generative models, such as DALL-E 3~\cite{betker2023improving} from OpenAI, Parti~\cite{yu2022scaling} from Google, and Stable Diffusion~\cite{rombach2022high} from Stability AI.}
These models allow for achieving a near-optimal point between reducing computing complexity and preserving visual details, greatly boosting the efficiency in both training and generation phase.
Among them, Stable Diffusion is the most  widely-used, which has already gained more than 10 million users \footnote{https://journal.everypixel.com/ai-image-statistics}.

\noindent
As latent generative models become more prevalent,  
the issues surrounding their potential for misuse are becoming increasingly important~\cite{schramowski2023safe,wang2023diagnosis,pan2024finding,liu2024latent,wen2023detecting,chen2023pathway}.
For example, malicious users may use the latent generative models to generate and distribute images containing inappropriate concepts such as ``\emph{sexual}'', ``\emph{drug use}'', ``\emph{weapons}'', and ``\emph{child abuse}''~\cite{schramowski2023safe}.
AI-powered plagiarism~\cite{francke2019potential} and IP (intellectual property) infringement problem surrounding the latent generative models are also important issues. For instance, users may synthesize high-quality images using 
one company's or open-sourced
latent generative models and then dishonestly present them as their own original artwork (e.g., photographs and paintings) to gain recognition and reputation, which is harmful to society and may cause a series of IP problems.
Consequently, it's crucial to be able to trace the source of images generated by latent generative models, i.e., determining if a certain image was produced by a specific model.

\noindent
There are several existing methods for tracing the source of the images.
Watermarking-based methods~\cite{luo2009reversible,pereira2000robust,tancik2020stegastamp} %
typically add watermark into the images and the images from specific origins can be identified via analyzing if the particular watermark is inserted in the images or not.
Classification-based approaches~\cite{sha2022fake} train multi-classes classifiers where each class corresponds to a specific origin (source model). Another set of methods inject fingerprints into the model during training~\cite{yu2019attributing,yu2021artificial} or by modifying the architectures of the models~\cite{yu2022responsible}, so that the images generated by the injected models will contain the fingerprinting and they can be detected by the fingerprinting decoders held by the model owner. All the above-listed methods share the limitation that 
requires extra steps during the training or generation phase, restricting their usage on the generated images without such operations. Also, many of the extra required operations might compromise the quality of the generated images. In addition, there is an increasing number of proposed attacks specifically targeting artificial watermarks, such as watermark stealing attacks~\cite{jovanovic2024watermark} and watermark forgery~\cite{wang2021watermark}. The usages of the artificial watermarks itself may also include the vulnerabilities.

\noindent
In this paper, we investigate whether it is possible to trace the images generated by a specific latent generative model without the aforementioned requirements such as adding artificial watermarks during the generation~\cite{tancik2020stegastamp,wen2023tree} and injecting fingerprintings during training~\cite{yu2021artificial,yu2022responsible,fernandez2023stable}. 
Inspired by ~\citet{wang2023origin}, we find that the input reverse-engineering based method is a promising way to achieve \textit{alteration-free origin attribution}.
Thus, we develop a latent inversion based method to trace the generated images of the inspected model by checking if the examined images can be well-reconstructed with an inverted latent input.  In detail, it works by reverse-engineering the latent input of the decoder in the inspected model for each examined image, and the examined image is considered as the generated image of the inspected model if the distance between the reconstructed image and the examined image is smaller than a pre-computed threshold. 

\begin{figure*}[]
	\centering
	\footnotesize
	\includegraphics[width=2.1\columnwidth]{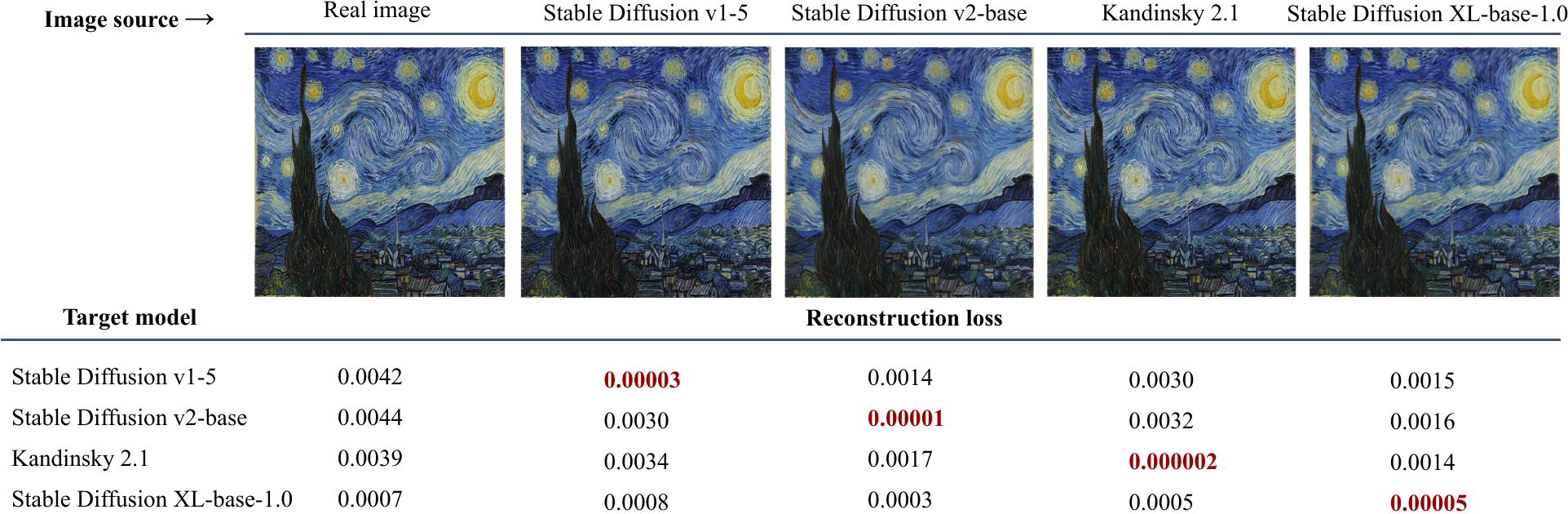}
 \vspace{-0.3cm}
 \caption{
     \textbf{Given a specific latent generative model,
    can we trace 
    the images generated by this model without extra artificial watermarks?} We show that 
    the images generated by the inspected model can be traced \emph{without any additional requirements on the model's training and generation phase (such as adding a watermark after generation~\cite{tancik2020stegastamp,wen2023tree} or 
    injecting fingerprinting during training~\cite{yu2021artificial,yu2022responsible,fernandez2023stable})} by using the reconstruction loss computed by our method.
    For example, here we show some images generated by different models and their reconstruction losses on different models. Even though the generated images from different models can look seemingly identical, their reconstruction loss can differ by an order of magnitude. The reconstruction loss is extremely low if the examined image is generated by the inspected model.}
 \label{fig:example_intro}
 \vspace{-0.6cm}
\end{figure*}

\noindent We observe that directly using the gradient-based optimization approach to invert the latent input suffers from \textit{sub-optimal effectiveness and low efficiency} on the state-of-the-art text-to-image latent generative models. We then find that the main reason for this phenomenon 
is the sub-optimal initialization (i.e., the randomly sampled starting point in the optimization process typically has a large distance to the ground-truth input). To solve this problem, we propose a strategy for finding a better starting point in optimization by exploiting the invertibility of the autoencoder, which is a main component in latent generative models. In particular, we leverage the latent projection of the given image by the encoder as an initialization. Our experiments on the state-of-the-art latent generative models (e.g., Stable Diffusion~\cite{rombach2022high} and Kandinsky~\cite{razzhigaev2023kandinsky}) show that our method is \textit{highly effective and more efficient} than existing methods for tracing the images generated by the inspected model in the alteration-free manner.
(\autoref{fig:example_intro} shows some examples).
Our results also reflect that 
the generated images of a specific model are naturally watermarked by the decoder module, which is essentially exploited by our method to enable alteration-free origin attribution.
\noindent
Our contributions are summarized as follows:
\ding{172} We propose a new alteration-free inversion-based 
origin attribution method (called \sys) designed for latent generative models which does not require additional operations
on the model’s training and generation phase. \ding{173} 
We evaluate our method on the state-of-the-art latent generative models, such as Stable Diffusion~\cite{rombach2022high} and Kandinsky~\cite{razzhigaev2023kandinsky}. The results show that our method is more \emph{effective and efficient} than existing alteration-free methods. As a result, our approach can even correctly identify the source from different versions of the Stable Diffusion models. \ding{174} We demonstrate the generalizibilty of our approach to  both diffusion and autoregressive models, and show effectiveness of our approach against both discrete and continuous autoencoders.
Our code can be found at \url{https://github.com/ZhentingWang/LatentTracer}.

\section{Related Work}
\label{sec:related}

\noindent
\textbf{Latent Generative Models.}
Significant progress in the image synthesis task has been made with the advent of latent generative models~\cite{rombach2022high,gu2022vector,luo2023latent}. 
The latent generative models leverage the features and visual patterns learned by the autoencoder, which can reduce the dimensionality of the samples in the generation process and help the models generate highly detailed images. Such latent-based architecture enable reaching a near-optimal balance between enhancing efficiency and preserving effectiveness.
A prime example of these models is the Stable Diffusion, which utilizes a diffusion process within a latent space derived from an autoencoder. 

\noindent
\textbf{Detection of AI-generated Images.} 
Detecting images generated by generative models (e.g., deepfake images~\cite{mirsky2021creation,wu2024traceevader}) has become more and more important due to the increasing concern about the potential misuse of these generated images~\cite{kietzmann2020deepfakes,flynn2021disrupting,flynn2022deepfakes,whittaker2020all,partadiredja2020ai}.
Many of the existing approaches~\cite{frank2020leveraging,dolhansky2020deepfake,wang2020cnn,zhao2021multi,corvi2023detection} frame this problem as a binary classification task, which aims at distinguishing between synthetic and authentic images.
These methods leverage the artifacts such as frequency signals~\cite{frank2020leveraging,durall2019unmasking,durall2020watch,jeong2022frepgan} and texture patterns~\cite{liu2020global} as the key features to solve this binary classification problem.
There are also existing works, such as DIRE~\cite{wang2023dire} which focus on differentiating between  real images and diffusion-generated images (i.e., images generated by all diffusion models).
Although these techniques have shown promise in identifying AI-generated images, they lack the ability to determine whether a specific image was created by a particular image generation model.

\noindent
\textbf{Origin Attribution of Generated Images.}
Many techniques have been developed to trace the origins of images, including watermark-based methods~\cite{swanson1996transparent,luo2009reversible,pereira2000robust,tancik2020stegastamp,wen2023tree}, classification-based approaches~\cite{sha2022fake}, and model fingerprinting methods~\cite{yu2019attributing,yu2021artificial,yu2022responsible,fernandez2023stable}. However, these methods are limited by the need for additional steps either in the training or image generation stages, which are not feasible for images produced without these processes. %
\textbf{In contrast, our method does not have such limitations.}
Recent research \citet{wang2023origin} shows that the
origin attribution without the requirements on additional steps can be achieved by reverse-engineering
the input for the model. 
However, it suffers from
sub-optimal effectiveness and low efficiency on the state-of-the-art latent generative models, especially in the cases for distinguishing the generated images of the inspected large latent generative model and that of other models having similar architectures. 
Another set of attribution methods~\cite{albright2019source,zhang2020attribution} focus on finding the source model of a specific image given a set of suspicious candidate models. These methods rely
on the assumptions that the inspector has the white-box access to every model in the candidate
set, and the inspected image must originate from one of these models. By contrast, our method does not have such assumptions.

\noindent
\textbf{Input Inversion for Generative Models.}
Previous works on generative model inversion techniques mainly focus on image editing applications~\cite{karras2020analyzing,jahanian2019steerability,zhu2016generative}. However, our focus is on a substantially different task - tracing the origins of generated images. Additionally, previous studies only include the studies on relatively outdated and small-scale GANs~\cite{Goodfellow2014GenerativeAN}. In contrast, we shift our attention to state-of-the-art large-scale latent generative models such as Stable Diffusion~\cite{rombach2022high}, which have more advanced architectures with significantly greater numbers of parameters, different training data distributions, and much larger training data diversity.

\section{Preliminary}
\label{sec:preliminary}

Inversion-based method is a promising way to achieve alteration-free origin attribution method~\cite{wang2023origin}.
In this section, we provide more background about it.
To facilitate discussion, we first introduce the definition of the belonging of image generation model.

\begin{definition}{(Belonging of Image Generation Model)}\label{def:membership}
    Given an image generation model \(\mathcal{M}: \mathcal{I} \mapsto \mathcal{X}_{\mathcal{M}}\), where \(\mathcal{I}\) denote the input space. \(\mathcal{X}_{\mathcal{M}}\) is the output space of the model. A sample \(\bm x\) is a \textbf{belonging} of model \(\mathcal{M}\) if and only if \(\bm x \in \mathcal{X}_{\mathcal{M}}\).
    It is a \textbf{non-belonging} if \(\bm x \notin \mathcal{X}_{\mathcal{M}}\).
\end{definition}

\noindent
Intuitively, the belongings of a latent generative model are the images generated by this model, and the non-belongings include the images generated by other models and the real images.
Then, we have the definition of the input inversion:

\begin{definition}{(Input Inversion)}\label{def:input_inversion}
Given an image generation model \(\mathcal{M}: \mathcal{I}
\mapsto \mathcal{X}\), where \(\mathcal{I}\) and \(\mathcal{X}\) are input space and pixel space of the image generation model, respectively, the latent inversion task for an image \(\bm x\) is finding the input \(\bm i^{\star}\) that makes the generated image \(\mathcal{M}(\bm i^{\star})\) as close as possible to the image \(\bm x\).
The reconstruction loss is defined as \(\mathcal{L}(\mathcal{M}(\bm i^{\star}),\bm x)\), where \(\mathcal{L}\) is a distance measurement.
\end{definition}

\noindent
Based on the definition of the input inversion, we have the definition of the inversion-based origin attribution:

\begin{definition}{(Inversion Based Origin Attribution)}\label{def:inversion_based_oa}
Given an inversion method \(\mathcal{R}: \mathcal{X} \mapsto \mathcal{I}\), the inversion-based origin attribution method determines a given sample \(\bm x\) is a belonging of a given model \(\mathcal M\) if the reconstruction loss of the reverse-engineered sample is smaller than a threshold value \(t\), i.e., \(\mathcal{L}(\mathcal{M}(\bm i^{\star}),\bm x) < t\).
\end{definition}

\noindent
The inversion-based origin attribution method will be highly effective if the reconstruction losses of the belonging images and that of the non-belonging images are well-separated by the threshold value.
The most straightforward way to conduct the inversion is using the gradient-based method~\cite{wang2023origin}, which is formally defined as follows:%

\begin{definition}{(Gradient-based Inversion)}\label{def:gradient}
Given a model \(\mathcal{M}\) and an image \(\bm x\), 
the gradient-based inversion searches the inverted input 
\(\bm i^{\prime}\) by repeatedly updating the input via the gradient on the reconstruction loss until converge. For each step, 
the searched input \(\bm i^{\prime}\) is updated via the following equation: \(\bm i^{\prime} = \bm i^{\prime} - lr\cdot \frac{\partial \mathcal{L}(\bm x, \mathcal{M}(\bm i^{\prime}))}{\partial \bm i^{\prime}}\), where \(lr\) is the learning rate and \(\mathcal{L}\) is the measurement of the distance.
\end{definition}

\section{Method}
\label{sec:method}

In this section, we first provide the formulation of the investigated problem
and then
introduce our method \sys designed for tracing the generated images of the inspected models.

\subsection{Problem Formulation}
\label{sec:pf}

\noindent
\textbf{Inspector's Goal.}
The goal of the inspector is tracing the belongings of the inspected model \(\mathcal M\) in an \emph{alteration-free manner (i.e., without any extra requirement during the developing or generation phase of the model)}.
It can be viewed as designing a tracing algorithm \(\mathcal B\) whose inputs are the examined image \(\bm x\) and the inspected model \(\mathcal M\), it then returns a flag representing whether the examined image is the belonging of the inspected model or not.
Formally, it can be written as \(\mathcal B: (\mathcal{M},\bm x) \mapsto \{0,1\}\), where 0 denotes image belonging to the model, and 1 denotes non-belonging. 

\noindent
\textbf{Inspector's Capability.}
The inspector has white-box access to the inspected model 
\(\mathcal{M}\), allowing for the access of intermediate outputs and the computation of the model's gradients. This assumption is practical, especially in the scenario where the inspector is the owner of the inspected model. However, the inspector cannot necessarily control the development and the generation process of the inspected model. Since our approach exploits the implicit watermarks in autoencoders module of latent generative models, we focus on traceability in latent generative models with unique and distinct auto-encoders. Distinguishing the belonging images of the inspected model and the images generated by other models sharing the same auto-encoder with the inspected model will be our future work.

\subsection{Latent Inversion}

\noindent
For the latent generative model \(\mathcal{M}\), the process for synthesizing images can be written as  
\(\bm x = \mathcal{M}(\bm p, \bm n) = \mathcal{D}(\mathcal{C}(\bm p, \bm n))\), where \(\bm p\) and \(\bm n\) are the conditional input and the unconditional noise for the latent generation process, respectively. 
\(\mathcal{D}\) is the decoder to transform the latent to the pixel space. \(\mathcal{C}\) here is the latent generation process in the latent space.
\(\bm x\) is the image synthesized by the model \(\mathcal{M}\).

\noindent \textbf{Asymmetric challenge of data generation and origin attribution.} A straightforward way to conduct the inversion-based origin attribution is by searching the input space of the whole model, including the conditional prompts and the unconditional noise, to identify whether an input would have led to generation of the given space. However, during the data generation phase, the users of the model have full control of selecting the 
hyper-parameters in the latent generation process (e.g., diffusion process and autoregressive process), such as the selection of different
diffusion samplers in latent diffusion models (e.g., DDIM~\cite{song2020denoising} and DPM-Solver~\cite{lu2022dpm}).
Therefore, inverting the input for the whole model is challenging as the inspector does not know the exact hyper-parameters used in the latent generation phase.

\noindent \textbf{Leveraging deterministic decoders for origin attribution.} In latent generative models, the first step is sampling a new latent vector from a latent generation process (e.g., diffusion process and autoregressive process), which is then upsampled to pixel space using a deterministic decoder. While the stochastic and asymmetric generation process in latent space creates an asymmetric challenge in traceability, as discussed above, we bypass it by solely focusing on traceability using the deterministic decoder.
As we discussed in \autoref{sec:pf}, we focus on the scenario where the latent generative models have unique autoencoders. In this case, if the examined image \(\bm x\) is generated by the decoder \(\mathcal{D}\), then it must be the belonging of the latent generative model \(\mathcal{M}\).
To design an origin attribution method that is \textit{orthogonal to the selection of the samplers and hyper-parameters during generation}, we convert the problem of origin attribution into detecting whether the images are generated by the decoder of the inspected model.

\begin{figure}[]
	\centering
	\footnotesize
	\includegraphics[width=0.8\columnwidth]{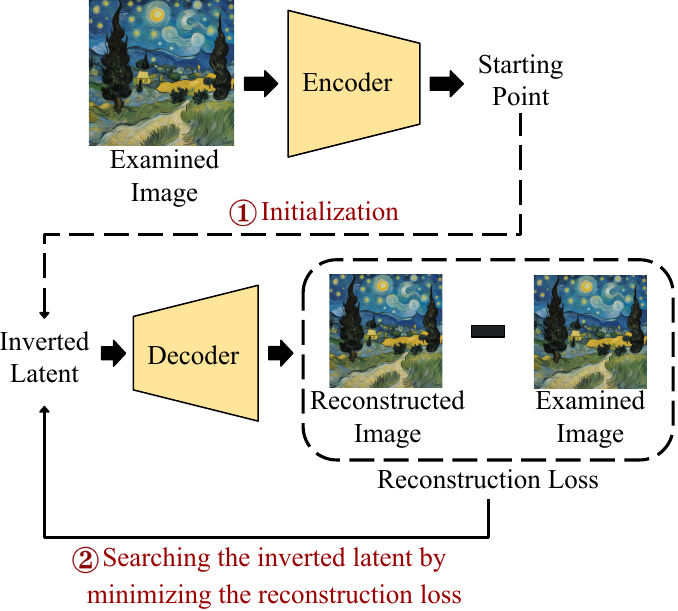}	
 \vspace{-0.3cm}
 \caption{Pipeline of our latent inversion method. 
 First, our method uses the corresponding encoder to get the starting point for the inversion. Then, it uses the gradient-based optimization to search the inverted latent by minimizing the reconstruction loss. The examined image is flagged as a belonging image of the inspected model if the final reconstruction loss is smaller than a threshold.
 }
 \label{fig:pipeline}
 \vspace{-0.4cm}
\end{figure}

\begin{figure*}[]
    \centering
    \footnotesize
    \begin{subfigure}[t]{0.65\columnwidth}
        \centering
        \footnotesize
        \includegraphics[width=\columnwidth]{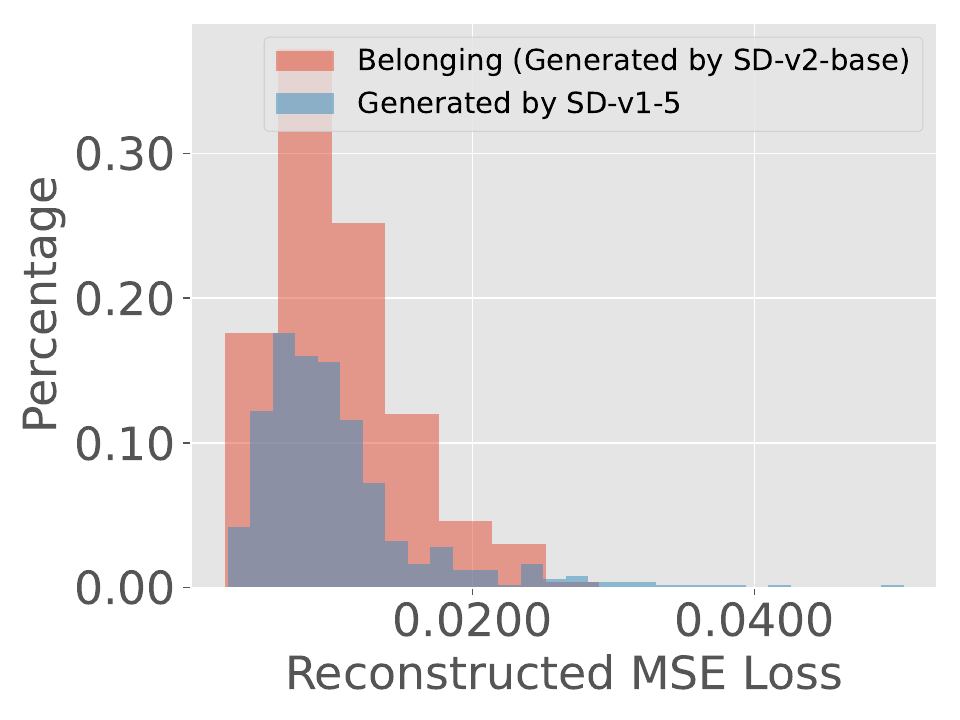}
        \caption{Random Initialization + Gradient-based Inversion~\cite{wang2023origin}}        \label{fig:distribution_encoderandgradient_random_init}
    \end{subfigure}
    \hfill
    \begin{subfigure}[t]{0.65\columnwidth}
        \centering
        \footnotesize
        \includegraphics[width=\columnwidth]{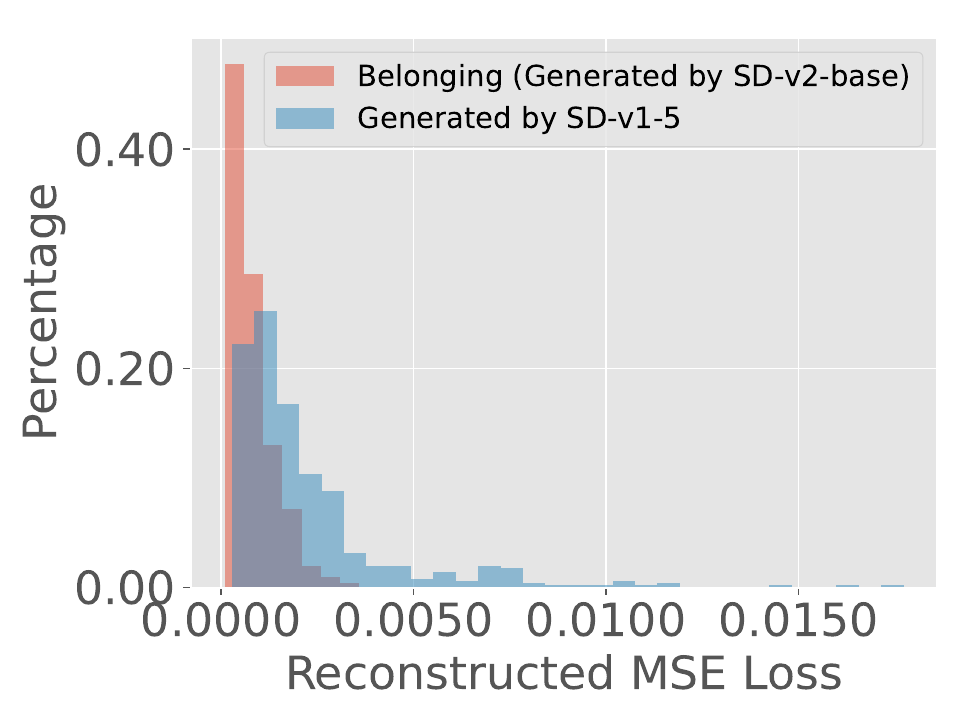}
        \caption{Encoder-based Inversion}        \label{fig:loss_distrubution_encoder}
    \end{subfigure}
    \hfill
    \begin{subfigure}[t]{0.65\columnwidth}
        \centering
        \footnotesize
        \includegraphics[width=\columnwidth]{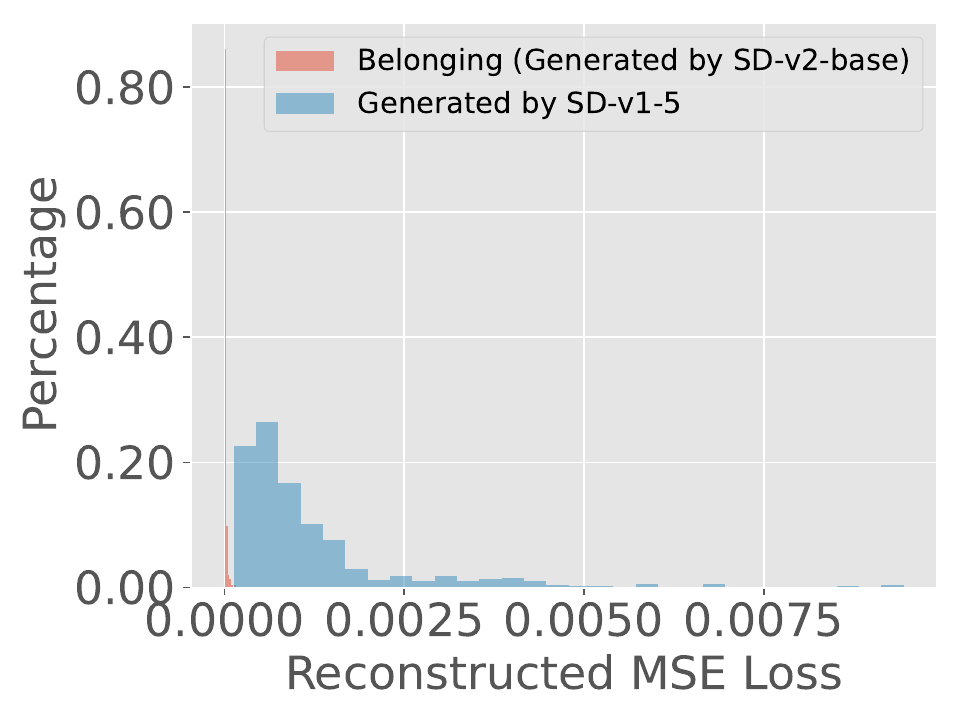}
        \caption{Encoder-based Initialization + Gradient-based Inversion (Ours)}
        \label{fig:distribution_encoderandgradient}
    \end{subfigure}
    \vspace{-0.2cm}
\caption{Comparison on the reconstruction loss distributions for different inversion methods.
The scenario is distinguishing the 500 images generated by the inspected model (i.e., Stable Diffusion v2-base) and the 500 images generated by other model (i.e., Stable Diffusion v1-5 here).
50 prompts sampled from PromptHero~\cite{prompthero} are used to generate these belonging images and non-belonging images (More details about the used prompts can be found in \autoref{sec:detailed_prompts}). Our method is highly effective since the reconstruction losses for the belongings and that for non-belongings are nearly completely separated in our method.
}\label{fig:compare}
\vspace{-0.4cm}
\end{figure*}
\begin{figure}[!b]
	\centering
	\footnotesize
        \vspace{-0.5cm}
	\includegraphics[width=0.7\columnwidth]{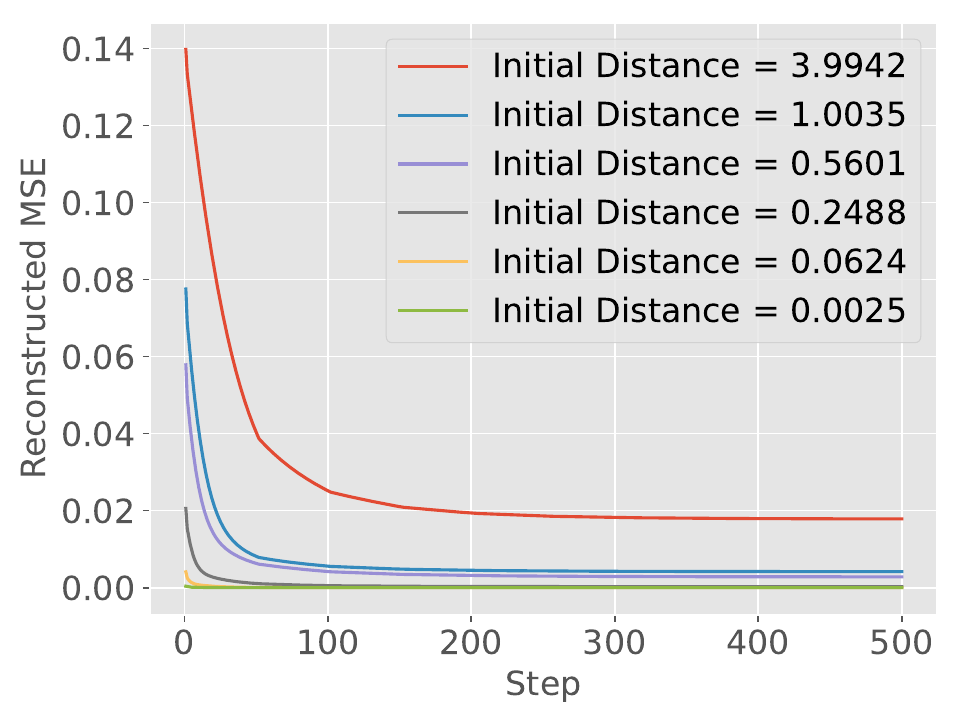}	
 \vspace{-0.3cm}
 \caption{Effects of the initial distance to the ground-truth latent. The starting 
point closer to the ground-truth latent will lead to better efficiency and effectiveness for the inversion.
 }\label{fig:init_dist2gt}
    \vspace{-0.6cm}
\end{figure}

Therefore, we focus on the \emph{latent inversion} introduced as follows: Given the decoder \(\mathcal{D}:\mathcal{A}
\mapsto \mathcal{X}\) in the latent generative model and the examined image \(\bm x\), where \(\mathcal{A}\) is the latent space and \(\mathcal{X}\) is the pixel space,
we aim at finding the latent \(\bm a^{\star}\) that makes the generated image \(\mathcal{D}(\bm a^{\star})\) as close as possible to the examined image \(\bm x\)
We then perform origin attribution based on the reconstruction loss of the latent inversion. In detail, we consider the examined image \(\bm x\) as a belonging if \(\mathcal{L}(\mathcal{D}(\bm a^{\star}),\bm x) < t\), where \(t\) is a threshold value.
The pipeline of our method is demonstrated in \autoref{fig:pipeline}.
Different from the inversion method in \citet{wang2023origin} that uses the random starting point for optimization, our method first uses the corresponding \textit{encoder} to obtain the starting point for the inversion. Following this, it
works by checking if the examined images can
be well-reconstructed on the inspected model with an inverted latent searched by the gradient optimization.
Our method is
generalizable to the models equipped with different types of auto-encoders, e.g., Continuous Auto-encoder~\cite{kingma2013auto} in Stable Diffusion~\cite{rombach2022high}, Quantized Auto-encoder~\cite{van2017neural}. Our method also works on Autoregressive Model~\cite{yu2021vector} used in Parti~\cite{yu2022scaling} (See \autoref{sec:different_ae}).
The design details of our method can be found in the following sections.

\subsection{Limitation of Canonical Gradient-based Inversion}
\label{sec:limitation_gradient}

\noindent
A straightforward way for conducting latent inversion is directly using the gradient-based method to search the inverted latent (similar to 
 \autoref{def:gradient}).
However, we find that directly using the gradient-based inversion 
has sub-optimal effectiveness and low efficiency on the state-of-the-art latent generative models, especially in the case of distinguishing the belonging images and the images generated by other models having similar architectures.
\autoref{fig:distribution_encoderandgradient_random_init} provides the empirical results indicating the sub-optimal effectiveness of the gradient-based latent inversion method since it demonstrates that the reconstruction losses of the belonging images and non-belonging images are not well-separated.
The experimental settings can be found in the caption of \autoref{fig:compare}. We also find that it requires large number of optimization steps to convergence.

\noindent
Upon investigation we find that a key factor for the efficiency and the effectiveness of the gradient-based latent inversion is the selection of the starting point of the optimization~\cite{bertsekas2009convex}.
More specifically, a starting point closer to the ground-truth latent (i.e., the latent used when generating the belonging image) will lead to better efficiency and effectiveness for the origin attribution. To confirm this, we create different starting points that have different initial distances to the ground-truth latent and collect their reconstruction losses during the optimization process. 
\autoref{fig:init_dist2gt} demonstrates that
a closer starting point will lead to a faster convergence speed and lower reconstruction loss after the convergence, and it confirms our analysis. \textbf{Technically, our contribution can be summarized as follows:} We observed the ineffectiveness and inefficiency for the existing inversion based origin attribution methods on the state-of-the-art large latent generative models, and
identify the key reason for this phenomenon. We propose a simple yet effective approach to solve this problem, and demonstrate that images produced by the state-of-the-art latent generative models may naturally carry an implicit watermark added by the decoder when decoding the latent samples as they can be detected by our method without any artificial watermarks.

\subsection{Our Approach: Exploiting the Invertibility of the Auto-encoder}

\noindent
In this section, we introduce our method \sys that combines the gradient-based method and the invertibility of the auto-encoder.
For the auto-encoder used in the latent generative models, the main training objective is to make the decoder have the ability to reconstruct the input of the encoder~\cite{kingma2013auto}, i.e., \(\bm x \approx \mathcal{D}(\mathcal{E}(\bm x))\), where \(\mathcal{E}\) is the encoder and \(\mathcal{D}\) is the decoder, \(\bm x\) denotes the image input. 
Existing image editing methods for the latent generative models~\cite{mokady2023null,parmar2023zero} demonstrate that the encoder (i.e., \(\mathcal{E}\)) of the auto-encoder used in the latent generative models
can also approximate the latent input of the decoder \(\mathcal{D}\), i.e., \(\bm a \approx \mathcal{E}(\mathcal{D}(\bm a))\), where \(\bm a\) is the input in the latent space.
Intuitively, we can directly use the encoder to invert the latent input of the decoder, and we define this inversion method as encoder-based inversion.

\begin{definition}{(Encoder-based Latent Inversion)}\label{def:encoder_based_inversion} 
Given a decoder \(\mathcal{D}\) and its corresponding encoder \(\mathcal{E}\), for an image \(\bm x\), 
the encoder-based latent inversion finds the inverted latent 
\(\bm a^{\prime}\) by directly exploiting the encoder, i.e., \(\bm a^{\prime} = \mathcal{E}(\bm x)\).
\end{definition}

\noindent
The encoder-based latent inversion is highly efficient since it only requires one forward process of the encoder and the run-time for it is nearly negligible.
The research question that we want to explore is: \emph{Is the encoder-based inversion effective for the origin attribution problem of the latent generative models?} 
We show results of a preliminary evaluation in \autoref{fig:loss_distrubution_encoder}, where the reconstruction loss of the belonging images (i.e., the images generated by the inspected model Stable Diffusion v2-base) is represented by red color, and the reconstruction losses of the images generated by other models are shown in the blue color. As can be seen, the reconstruction losses of the belonging samples and those of the non-belonging samples are not well-separated. Thus, we have the following observation:

\begin{observation}\label{def:ob_1}
Encoder-based latent inversion method has low effectiveness for the origin attribution problem.
\end{observation}

\begin{table}[]
\centering
\scriptsize
\setlength\tabcolsep{3pt}
\caption{Comparison for the initial distances
from the ground-truth latent to the starting points generated by different initialization methods. The reported number is the mean and standard deviation values of 100 different belonging samples.}\label{tab:init_dist}
\begin{tabular}{@{}ccc@{}}
\toprule
\multirow{3}{*}{Model}   & \multicolumn{2}{c}{Initial Distance} \\ \cmidrule(l){2-3} 
                         & \makecell{Random\\ Initialization} & \makecell{Encoder-based\\ Initialization} \\ \midrule
Stable Diffusion v1-5    & 1.930$\pm$0.021      & 0.019$\pm$0.007       \\
Stable Diffusion v2-base & 1.844$\pm$0.017      & 0.014$\pm$0.004       \\ \bottomrule
\end{tabular}
\end{table}

\begin{figure}[]
    \centering
    \footnotesize
    \begin{subfigure}[t]{0.48\columnwidth}
        \centering
        \footnotesize
        \includegraphics[width=\columnwidth]{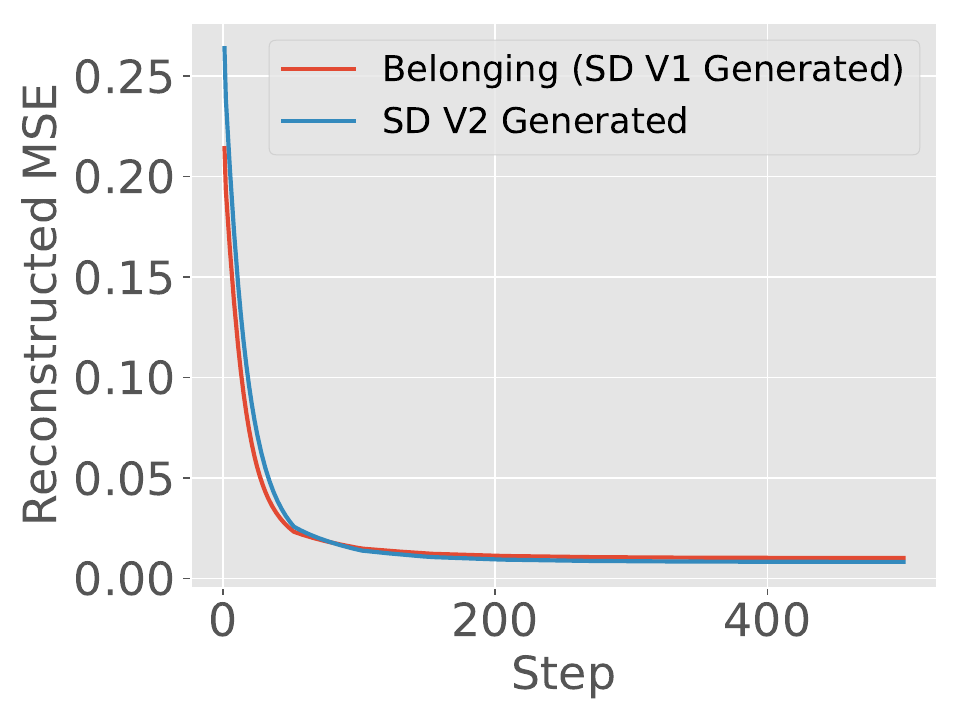}
        \vspace{-0.6cm}
        \caption{Random-based}        \label{fig:hidden_trigger}
    \end{subfigure}
    \begin{subfigure}[t]{0.48\columnwidth}
        \centering
        \footnotesize
        \includegraphics[width=\columnwidth]{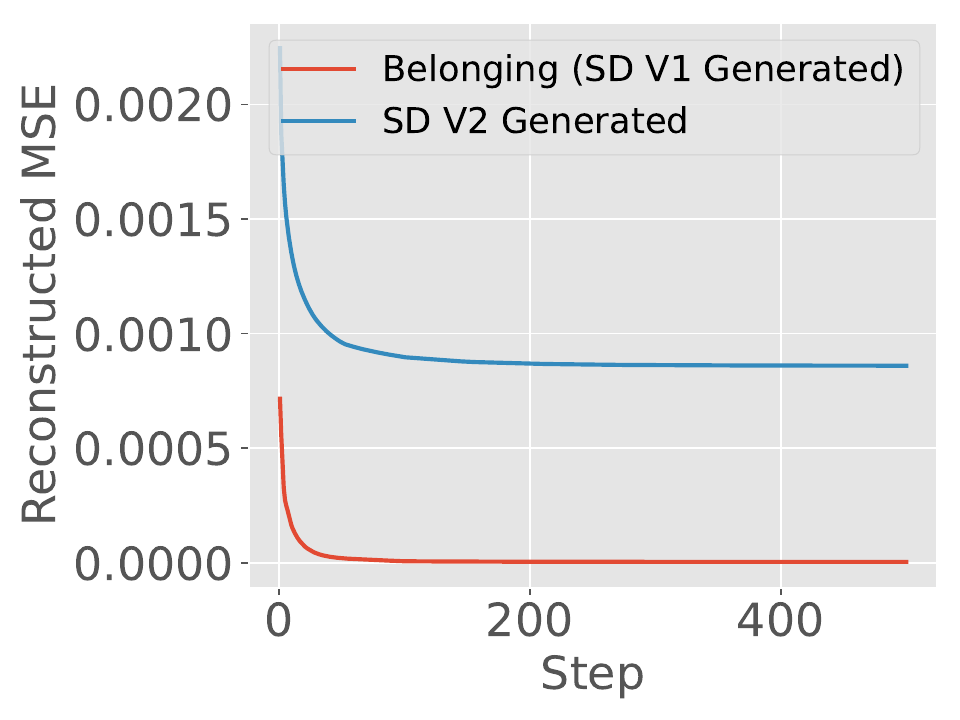}
        \vspace{-0.6cm}
        \caption{Encoder-based}        \label{fig:compare_random_encoder_losslist_encoder3}
    \end{subfigure}
\caption{Comparison on the reconstruction loss curve for the random starting point and the encoder-generated starting point. The encoder-based starting point lead to a faster convergence speed and a better separation of the reconstruction losses.}\label{fig:loss_curve}
\vspace{-0.3cm}
\end{figure}

\noindent
This observation is expected since the auto-encoder only provides an approximation of the inversion.
Although directly using the encoder to invert the input in the latent space leads to low effectiveness, we still can exploit the invertibility of the auto-encoder. Note that in \autoref{sec:limitation_gradient}, we discussed that the effectiveness and the efficiency of the gradient-based inversion are highly sensitive to the starting point before the optimization. Since the auto-encoder used in the latent generative models has potential invertibility, another question we want to explore is: \emph{Before the optimization process of the gradient-based latent inversion, is it possible to use the encoder to get the starting point that is better than the random starting point?}
To investigate this question, we compare the distance from the randomly generated starting point to the ground-truth latent and that from the encoder-generated starting point. The results are shown in \autoref{tab:init_dist}.
The models used here are the Stable Diffusion v1-5 and the Stable Diffusion v2-base. 
The results indicate that the encoder-generated starting point's distance to the 
ground-truth latent is much smaller than that of the random-generated starting point.
To further explore the effectiveness and the efficiency of the encoder-based latent initialization with the gradient-based inversion method, we record the reconstruction loss curve for both the random latent initialization with the gradient-based inversion and the encoder-based latent initialization with the gradient-based inversion. The inspected model here is the Stable Diffusion v1-5. The belonging image and the non-belonging image here are generated by using the same prompt.
As shown in \autoref{fig:loss_curve}, the convergence speed of using the encoder-generated starting point is much faster than that of using the random starting point. At the same time, the encoder-based initialization leads to an obvious separation of the reconstructed losses on belonging samples and the non-belonging samples, while the random initialization does not have such an effect.
The results for the reconstruction losses of 500 belonging images and 500 non-belonging images in \autoref{fig:distribution_encoderandgradient_random_init} and \autoref{fig:distribution_encoderandgradient} also show that the gradient-based inversion with encoder-based initialization is much more effective than the inversion with random initialization since the former leads to a much higher separability for the reconstruction losses of the belonging samples and the non-belonging samples.
Based on these results, we have the following observation:

\begin{observation}\label{def:ob_2}
For gradient-based latent inversion and origin attribution, the encoder-based initialization leads to better effectiveness and efficiency compared to the random initialization.
\end{observation}

\noindent
\textbf{Algorithm.} We design our algorithm based on our analysis and observations.
Our algorithm, detailed in \autoref{alg:detection}, takes as input the image under examination, denoted as 
\(\bm x\), alongside the inspected model 
\(\mathcal{M}\). Its output manifests as the inference outcomes, determining whether the examined image either is the belonging of the inspected model or not.
At the first step of the process, the algorithm
determines the threshold for the detection.
In detail, it
utilizes the model 
\(\mathcal{M}\)
 to generate 
\(N\) (defaulting to 100 in this paper) images from randomly selected prompts. It then proceeds to calculate the mean value (\(\mu\)) and standard deviation (\(\sigma\)) of the reconstruction loss on the belonging samples of the model.
Following \citet{wang2023origin}, we use Grubbs' Hypothesis Testing~\cite{grubbs1950sample} to determine the threshold:

\begin{algorithm}[t]
 	\caption{Origin Attribution by Gradient-based Inversion with Encoder-based Initialization}\label{alg:detection}
    {\bf Input:} %
    \hspace*{0.05in} Model: \(\mathcal{M}\), Examined Data: \(\bm x\)\\
    {\bf Output:} %
    \hspace*{0.05in} Inference Results: Belonging or Non-belonging
	\begin{algorithmic}[1]
	     \Function {Inference}{$\mathcal{M}, \bm x$}
      \State \(\mathcal{E} = \mathcal{M}\).Encoder
      \State \(\mathcal{D} = \mathcal{M}\).Decoder
      \LineComment{Obtaining Threshold (Offline)}
      \State \(t \leftarrow\) Calculating Threshold 
 [\autoref{eq:hypo}]%
      \LineComment{Reverse-engineering}
      \State \(\bm a = \mathcal{E}(\bm x)\)
      \For{\(e \leq \rm{max\_epoch}\)}
      \State \(cost = \mathcal{L}\left(\mathcal{D}(\bm a), \bm x\right)\) 
      \State \(\Delta_{\bm a} = \frac{\partial cost}{\partial \bm a}\) 
      \State \(\bm a = \bm a - lr\cdot \Delta_{\bm a}\) 
      \EndFor

      \LineComment{Determining Belonging}
      \If{ \( cost \leq t \) }
      \State \Return{$\rm Belonging$}
      \Else
      \State \Return{$\rm NonBelonging$}
      \EndIf 
      
    \EndFunction
    \end{algorithmic}
\end{algorithm}

\vspace{-0.4cm}
\begin{equation}
\label{eq:hypo}
t = \frac{(N-1)\sigma}{\sqrt{N}} \sqrt{\frac{\left(t_{\alpha /N, N-2}\right)^2}{N-2+\left(t_{\alpha /N, N-2}\right)^2}} + \mu
\end{equation}
\vspace{-0.4cm}

\noindent
Here \(t_{\alpha /N, N-2}\) is the critical value of the \(t\) distribution with  \(N-2\) degrees of freedom and a significance level of \(\alpha /N\), where \(\alpha\) is the significance level of the hypothesis testing (i.e., 0.05 by default in this
paper). More details of the critical value can be found in \autoref{sec:appendix_critical}.
This calculation, being an offline process, necessitates execution only once per model.
In Line 7, we calculate the starting point of the optimization by exploiting the forward process of the encoder.
Moving on to Line 8-11, the reconstruction loss is calculated and the
inverted latent is optimized by gradient descent optimizer.
Note that the starting point is obtained by using the encoder \(\mathcal E\) to encode the inspected image \(\bm x\).
Lastly, Line 13-16 involves determining the affiliation of the examined data 
\(\bm x\) with the model 
\(\mathcal{M}\).

\section{Experiments and Results}
\label{sec:eval}

We first introduce the experiment setup (\autoref{sec:eval_setup}). We then evaluate the effectiveness (\autoref{sec:eval_effectiveness}) and the efficiency (\autoref{sec:eval_efficiency}) of our method \sys.
We also study the results on different types of auto-encoders in \autoref{sec:different_ae} and compare our method to existing methods requiring extra operations during training phase or generation phase in \autoref{sec:compare_perturbation}.

\subsection{Experiment Setup}
\label{sec:eval_setup}

Our method is implemented with Python 3.10 and PyTorch 2.0.
We conducted all experiments on a Ubuntu 20.04 server equipped with 8 A100 GPUs (one experiment/GPU).

\noindent
\textbf{Models.}
There are six state-of-the-art latent generative models involved in the experiments: Stable Diffusion v1-4
(SD v1-4), Stable Diffusion v1-5
(SD v1-5), Stable Diffusion v2-base
(SD v2-base), Stable Diffusion v2-1
(SD v2-1), Stable Diffusion XL-1.0-base
(SD XL-1.0-base) and Kandinsky 2.1.
Details of the models are in \autoref{sec:appendix_model_details}.

\noindent
\textbf{Evaluation Metrics.}
The effectiveness of the origin attribution methods is measured by calculating the accuracy of detection (Acc).
For an inspected model, when dealing with a mix of images that are either belonging or non-belonging of it, Acc represents the proportion of accurately identified images to the total number of images.
Additionally, we present a comprehensive count of True Positives (TP, or correctly identified belonging images), False Positives (FP, or non-belongings identified as belongings), False Negatives (FN, or belongings identified as non-belongings), and True Negatives (TN, or non-belongings correctly identified).

\begin{table*}[]
\centering
\scriptsize
\setlength\tabcolsep{4pt}
\caption{Results for distinguishing belonging images and images generated by other models. Here, Model \(\mathcal{M}_1\) is the inspected model, Model \(\mathcal{M}_2\) is the other model. As we discussed in \autoref{sec:pf}, we focus on the traceability in latent generative models with unique and
distinct auto-encoders. Thus, we do not include the setting where \(\mathcal{M}_1\)
and
\(\mathcal{M}_2\) share the same autoencoder, e.g.,  \(\mathcal{M}_1\)=SD v1-4 and \(\mathcal{M}_2\)=SD v1-5. To our understanding, RONAN~\cite{wang2023origin} is the only method that shares the same problem formulation as our method.}\label{tab:different_latent_archs}
\begin{tabular}{@{}ccccccccccccc@{}}
\toprule
\multirow{2}{*}{Model \(\mathcal{M}_1\)} & \multirow{2}{*}{Model \(\mathcal{M}_2\)} & \multicolumn{5}{c}{RONAN}                        &  & \multicolumn{5}{c}{\sys (Ours)}     \\ \cmidrule(lr){3-7} \cmidrule(l){9-13} 
                                         &                                          & TP                      & FP  & FN & TN & Acc    &  & TP  & FP & FN & TN  & Acc    \\ \midrule
\multirow{4}{*}{SD v1-4}                 & SD v2-base                               & 477                     & 485 & 23 & 15 & 49.2\% &  & 480 & 53 & 20 & 447 & 92.7\% \\
                                         & SD v2-1                                  & 480                     & 485 & 20 & 15 & 49.5\% &  & 482 & 55 & 18 & 445 & 92.7\% \\
                                         & SD XL-1.0-base                           & 475                     & 470 & 25 & 30 & 50.5\% &  & 476 & 81 & 24 & 419 & 89.5\% \\
                                         & Kandinsky                                & 484                     & 470 & 16 & 30 & 51.4\% &  & 475 & 65 & 25 & 435 & 91.0\% \\ \midrule
\multirow{4}{*}{SD v1-5}                 & SD v2-base                               & 477                     & 484 & 23 & 16 & 49.3\% &  & 481 & 55 & 19 & 445 & 92.6\% \\
                                         & SD v2-1                                  & \multicolumn{1}{l}{478} & 482 & 22 & 18 & 49.6\% &  & 479 & 54 & 21 & 444 & 92.5\% \\
                                         & SD XL-1.0-base                           & \multicolumn{1}{l}{477} & 472 & 23 & 28 & 50.5\% &  & 477 & 85 & 23 & 415 & 89.2\% \\
                                         & Kandinsky                                & 482                     & 469 & 18 & 31 & 51.3\% &  & 477 & 66 & 23 & 434 & 91.1\% \\ \midrule
\multirow{4}{*}{SD v2-base}              & SD v1-4                                  & 481                     & 425 & 19 & 75 & 55.6\% &  & 482 & 0  & 18 & 500 & 98.2\% \\
                                         & SD v1-5                                  & \multicolumn{1}{l}{480} & 422 & 20 & 78 & 55.8\% &  & 480 & 0  & 20 & 500 & 98.0\% \\
                                         & SD XL-1.0-base                           & 476                     & 480 & 24 & 20 & 49.6\% &  & 480 & 53 & 20 & 447 & 92.7\% \\
                                         & Kandinsky                                & 473                     & 473 & 27 & 27 & 50.0\% &  & 478 & 1  & 22 & 499 & 97.7\% \\ \midrule
\multirow{5}{*}{Kandinsky}               & SD v1-4                                  & 481                     & 479 & 19 & 21 & 50.2\% &  & 474 & 12 & 26 & 488 & 96.2\% \\
                                         & SD v1-5                                  & 483                     & 480 & 17 & 20 & 50.3\% &  & 476 & 11 & 24 & 489 & 97.5\% \\
                                         & SD v2-base                               & 482                     & 478 & 18 & 22 & 50.4\% &  & 480 & 46 & 20 & 454 & 93.4\% \\
                                         & SD v2-1                                  & \multicolumn{1}{l}{483} & 475 & 17 & 25 & 50.8\% &  & 479 & 46 & 21 & 454 & 93.3\% \\
                                         & SD XL-1.0-base                           & 478                     & 490 & 22 & 10 & 48.8\% &  & 480 & 82 & 20 & 418 & 89.8\% \\ \bottomrule
\end{tabular}
\vspace{-0.4cm}
\end{table*}

\subsection{Effectiveness}
\label{sec:eval_effectiveness}
In this section, we study the effectiveness of our method on the origin attribution task. 
We first investigate the effectiveness on distinguishing belonging images and images generated by other models, following with the study for the effectiveness on distinguishing belongings and real images.

\noindent
\textbf{Distinguishing Belonging Images and Images Generated by Other Models.}
In this section, we study our method's effectiveness on distinguishing between belonging images of a particular model and the images generated by other models. 
To measure the effectiveness of our method,
we first randomly collect 50 prompts designed for the state-of-the-art text-to-image generative models on the PromptHero~\cite{prompthero} (a main-stream website for prompt engineering of the generative models). For each prompt, we generate 10 samples on each model by using different random seeds. Thus, we have 500 generated samples for each model. For different inspected models, we use our method to distinguish the belonging images and the images generated by other models.
We compare our method to the existing reverse-engineering based origin attribution method RONAN~\cite{wang2023origin}. 
While most of the existing methods (such as image watermarking~\cite{wen2023tree} and model finger-printing~\cite{yu2022responsible}) require extra operations during the training or generation phase, RONAN is the only method that can achieve alteration-free origin attribution and \emph{has the same threat model with our method}.
The results are shown in \autoref{tab:different_latent_archs}, where Model \(\mathcal{M}_1\) denotes the inspected model, and Model \(\mathcal{M}_2\) represents the other model.
For our method, the average detection accuracy (Acc) of our method is 93.4\%, confirming its good performance for distinguishing between the belongings of the inspected model and the images generated by other models. 
As can be seen, the detection accuracy of RONAN is only around 50\% in our setting.
The results indicate our method outperforms the existing alteration-free origin attribution 
method by a large margin on the origin attribution for the state-of-the-art latent generative models. In \autoref{sec:compare_perturbation}, we also compare our method to existing state-of-the-art methods that require extra operations during the training or generation phase.

\noindent
\textbf{Distinguishing Belonging Images and Real Images.}
In this section, we investigate our approach's effectiveness in distinguishing between belonging images of a particular model and real images. The investigated models 
include Stable Diffusion v1-4, Stable Diffusion v1-5, Stable Diffusion v2-base, and Kandinsky.
We first randomly sample 500 images and their corresponding text captions in the LAION dataset~\cite{schuhmann2022laion}, which is the training dataset for many state-of-the-art text-to-image latent generative models such as Stable Diffusion~\cite{rombach2022high}.
For each model, we use the randomly sampled text captions as the prompts to generate 500 images as the belonging images.
The randomly sampled images from the LAION dataset are used as the non-belonging samples here.
The results are demonstrated in \autoref{tab:belonging_vs_trainingsamples}. 
On average, the detection accuracy (Acc) is 97.9\%. The results show that our method achieves good performance in distinguishing belonging images of a particular model and real images.

\begin{table}[]
\centering
\scriptsize
\setlength\tabcolsep{4pt}
\caption{Results for distinguishing belongings and real images. The real images used are randomly sampled from LAION~\cite{schuhmann2022laion}.}\label{tab:belonging_vs_trainingsamples}
\begin{tabular}{@{}cccccc@{}}
\toprule
Inspected Model  & TP  & FP & FN & TN  & Acc    \\ \midrule
SD v1-4    & 484 & 10 & 16 & 490 & 97.4\% \\
SD v1-5    & 487 & 9  & 13 & 491 & 97.8\% \\
SD v2-base & 487 & 6  & 13 & 494 & 98.1\% \\
Kandinsky                & 483 & 0  & 17 & 500 & 98.3\% \\ \bottomrule
\end{tabular}
\vspace{-0.3cm}
\end{table}

\subsection{Efficiency}
\label{sec:eval_efficiency}

In this section, we study the efficiency of our method. We collect the runtime for inferring if an examined image is the belonging of the inspected model. Four models (i.e., Stable Diffusion v1-4, Stable Diffusion v1-5, Stable Diffusion v2-base, Kandinsky) are included in this section as the inspected models. For each model, we run 5 times and the average runtime is reported in \autoref{tab:time}. Besides the runtime for our method (gradient-based inversion with encoder-based initialization), we also show the average runtime for the gradient-based inversion with random initialization. As can be observed, the speed of our method is much faster than the method with random initialization. This is because the starting point obtained by the encoder-based initialization is much closer to the ground-truth, leading to a faster convergence speed. 
Based on our experiments, our method only requires 100 optimization steps to converge, while the method with random initialization needs 400 steps (also see \autoref{fig:loss_curve}). In conclusion, our method is much more efficient than \citet{wang2023origin}.

\begin{table}[]
\centering
\scriptsize
\setlength\tabcolsep{3pt}
\caption{Average runtime on different models.}\label{tab:time}
\begin{tabular}{@{}cccc@{}}
\toprule
\multirow{2}{*}{Inspected Model} & \multicolumn{3}{c}{Runtime}                           \\ \cmidrule(l){2-4} 
                                 & \makecell{with Random\\ Initialization} &  & \makecell{with Encoder-based\\ Initialization (Ours)} \\ \midrule
SD v1-4            & 87.6s                &  & \textbf{21.5s}                       \\
SD v1-5            & 88.1s                &  & \textbf{21.8s}                       \\
SD v2-base         & 93.5s                &  & \textbf{23.2s}                       \\
Kandinsky                        & 120.7s               &  & \textbf{30.3s}                       \\ \bottomrule
\end{tabular}
\vspace{-0.2cm}
\end{table}

\subsection{Results on Different Types of Models}
\label{sec:different_ae}

\noindent
It is possible that the latent generative models may use different types of auto-encoders or different latent generation processes (e.g., diffusion and autoregressive process).
In this section, we discuss our method's generalization to different types of models. To study this, we conduct experiments on two types of autoencoders (i.e., Continuous Auto-encoder and Quantized Auto-encoder). Besides the experiments on the models with diffusion-based latent generation process (\autoref{sec:eval_effectiveness}), we also evaluate our method on Autoregressive Model in this section.
For Continuous Auto-encoder, we use the VAE~\cite{kingma2013auto} model used in the Stable Diffusion v1-5. For Quantized Auto-encoder, we use a VQ-VAE~\cite{van2017neural} model trained on the CIFAR-10 dataset~\cite{krizhevsky2009learning}. For Autoregressive Model, we use a ViT-VQGAN~\cite{yu2021vector} model trained on the ImageNet dataset~\cite{russakovsky2015imagenet}.
We randomly sample 1000 images from the LAION~\cite{schuhmann2022laion}, CIFAR-10~\cite{krizhevsky2009learning}, and ImageNet~\cite{russakovsky2015imagenet} dataset as the non-belonging images for the VAE, VQ-VAE, and ViT-VQGAN, respectively.
To obtain the belonging images that share similar complexity to these real images, we then use these models to conduct reconstruction on these real images and obtain the belonging images of the decoders that are similar to these real images, i.e., \(\bm x^{\prime} = \mathcal{D}(\mathcal{E}(\bm x))\) where \(\bm x\) is the real images and \(\bm x^{\prime}\) is the obtained belonging images similar to the corresponding real images. \(\mathcal{E}\) and \(\mathcal{D}\) are the encoder and the decoder, respectively.
After we obtain all the belonging images and non-belonging images for each model, we use our method to distinguish them (\autoref{tab:different_ae}) and find that our method has above 97\% detection accuracy on all different types of auto-encoders or latent generation processes, showing our method's generalization ability.

\begin{table}[]
\centering
\scriptsize
\setlength\tabcolsep{3pt}
\caption{Generalization to different types of models.}\label{tab:different_ae}
\begin{tabular}{@{}ccc@{}}
\toprule
Type                    & Model                                               & Acc    \\ \midrule
Continuous Auto-encoder & VAE~\cite{kingma2013auto}     & 98.2\% \\
Quantized Auto-encoder  & VQ-VAE~\cite{van2017neural}   & 97.5\% \\
Autoregressive Model    & ViT-VQGAN~\cite{yu2021vector} & 98.4\% \\ \bottomrule
\end{tabular}
\vspace{-0.2cm}
\end{table}

\subsection{Comparison to Methods Requiring Extra Steps}
\label{sec:compare_perturbation}

In this section, we provide the comparison to more state-of-the-art methods that require extra operations during the training or generation phase to illustrate a broader spectrum of comparison. The model used here is Latent Diffusion Model~\cite{rombach2022high}, and the dataset here is MS-COCO. The results are shown in \autoref{tab:compare_perturbation}. We report the SSIM value and the FID value between the original samples and the watermarked/fingerprinted generated samples. 
For SSIM (Structural Similarity Index Measure), a value of 1 indicates perfect similarity between two compared images, meaning they are exactly the same.
For FID (Fréchet Inception Distance), a value of 0 indicates that the two distributions are identical.
Since our method \sys does not have any watermarking/fingerprinting process, we can view the watermarked/fingerprinted 
samples of our method are identical to the original generated samples. Consequently, our method achieves a perfect SSIM score of 1 and an FID score of 0.
As can be observed, the methods necessitating extra steps have non-negligible negative influences on the generation quality, while our \sys guarantees no quality degradation.

\begin{table}[]
\centering
\scriptsize
\setlength\tabcolsep{3pt}
\caption{Comparison to methods requiring extra steps.}\label{tab:compare_perturbation}
\begin{tabular}{@{}ccc@{}}
\toprule
Method                           & SSIM & FID  \\ \midrule
Dct-Dwt~\cite{cox2007digital}            & 0.97 & 19.5 \\
SSL Watermark~\cite{fernandez2022watermarking} & 0.86 & 20.6 \\
FNNS~\cite{kishore2021fixed}             & 0.90 & 19.0 \\
HiDDeN~\cite{zhu2018hidden}             & 0.88 & 19.7 \\
Stable Signature~\cite{fernandez2023stable} & 0.89 & 19.6 \\
\sys (Ours)                             & 1.00 & 0    \\ \bottomrule
\end{tabular}
\vspace{-0.2cm}
\end{table}

\section{Conclusion}
\label{sec:conclusion}
We propose a latent inversion based method to detect the generated images of the inspected model by checking if the examined images can be well-reconstructed with an inverted latent input. Our experiments on different latent generative models demonstrate that our approach is highly accurate in differentiating between images produced by the inspected model and other images. 
Our results also imply an interesting direction that images created by today's latent generative models may inherently carry an implicit watermark added by the decoder when decoding the latent samples.

\section{Potential Broader Impact }
Studying the security and privacy aspects of the machine learning models potentially has ethical concerns~\cite{carlini2023extracting,carlini2023poisoning,zou2023universal,wang2022bppattack,tao2022backdoor}. In this paper, we design a method for tracing the generated images of a specific inspected latent generative model. We are confident that our method will protect the intellectual property and
improve the security of latent generative models and will be advantageous for the development of responsible AI~\cite{gu2024responsible,li2023self,guo2024domain,huang2023can,wang2023free,shao2024explanation,li2023black,wang2022unicorn}.

\section*{Acknowledgement}
We thank the anonymous reviewers for their valuable comments.
This research is supported by Sony AI, IARPA TrojAI W911NF-19-S-0012, NSF 2342250 and 2319944.
It is also partially funded by research grants to D. Metaxas through NSF 2310966, 2235405, 2212301, 2003874, and AFOSR-835531.
Any opinions, findings, and conclusions expressed in this paper are those of the authors only and do not necessarily reflect the views of any funding agencies.

\clearpage
\bibliographystyle{icml2024}
\bibliography{reference}

\begin{thebibliography}{82}
\providecommand{\natexlab}[1]{#1}
\providecommand{\url}[1]{\texttt{#1}}
\expandafter\ifx\csname urlstyle\endcsname\relax
  \providecommand{\doi}[1]{doi: #1}\else
  \providecommand{\doi}{doi: \begingroup \urlstyle{rm}\Url}\fi

\bibitem[Albright \& McCloskey(2019)Albright and McCloskey]{albright2019source}
Albright, M. and McCloskey, S.
\newblock Source generator attribution via inversion.
\newblock In \emph{CVPR Workshops}, volume~8, pp.\ ~3, 2019.

\bibitem[Bertsekas(2009)]{bertsekas2009convex}
Bertsekas, D.
\newblock \emph{Convex optimization theory}, volume~1.
\newblock Athena Scientific, 2009.

\bibitem[Betker et~al.(2023)Betker, Goh, Jing, Brooks, Wang, Li, Ouyang, Zhuang, Lee, Guo, et~al.]{betker2023improving}
Betker, J., Goh, G., Jing, L., Brooks, T., Wang, J., Li, L., Ouyang, L., Zhuang, J., Lee, J., Guo, Y., et~al.
\newblock Improving image generation with better captions.
\newblock \emph{Computer Science. https://cdn. openai. com/papers/dall-e-3. pdf}, 2:\penalty0 3, 2023.

\bibitem[Carlini et~al.(2023{\natexlab{a}})Carlini, Hayes, Nasr, Jagielski, Sehwag, Tramer, Balle, Ippolito, and Wallace]{carlini2023extracting}
Carlini, N., Hayes, J., Nasr, M., Jagielski, M., Sehwag, V., Tramer, F., Balle, B., Ippolito, D., and Wallace, E.
\newblock Extracting training data from diffusion models.
\newblock \emph{arXiv preprint arXiv:2301.13188}, 2023{\natexlab{a}}.

\bibitem[Carlini et~al.(2023{\natexlab{b}})Carlini, Jagielski, Choquette-Choo, Paleka, Pearce, Anderson, Terzis, Thomas, and Tram{\`e}r]{carlini2023poisoning}
Carlini, N., Jagielski, M., Choquette-Choo, C.~A., Paleka, D., Pearce, W., Anderson, H., Terzis, A., Thomas, K., and Tram{\`e}r, F.
\newblock Poisoning web-scale training datasets is practical.
\newblock \emph{arXiv preprint arXiv:2302.10149}, 2023{\natexlab{b}}.

\bibitem[Chen et~al.(2023)Chen, Fu, and Lyu]{chen2023pathway}
Chen, C., Fu, J., and Lyu, L.
\newblock A pathway towards responsible ai generated content.
\newblock \emph{arXiv preprint arXiv:2303.01325}, 2023.

\bibitem[Corvi et~al.(2023)Corvi, Cozzolino, Zingarini, Poggi, Nagano, and Verdoliva]{corvi2023detection}
Corvi, R., Cozzolino, D., Zingarini, G., Poggi, G., Nagano, K., and Verdoliva, L.
\newblock On the detection of synthetic images generated by diffusion models.
\newblock In \emph{ICASSP 2023-2023 IEEE International Conference on Acoustics, Speech and Signal Processing (ICASSP)}, pp.\  1--5. IEEE, 2023.

\bibitem[Cox et~al.(2007)Cox, Miller, Bloom, Fridrich, and Kalker]{cox2007digital}
Cox, I., Miller, M., Bloom, J., Fridrich, J., and Kalker, T.
\newblock \emph{Digital watermarking and steganography}.
\newblock Morgan kaufmann, 2007.

\bibitem[Dolhansky et~al.(2020)Dolhansky, Bitton, Pflaum, Lu, Howes, Wang, and Ferrer]{dolhansky2020deepfake}
Dolhansky, B., Bitton, J., Pflaum, B., Lu, J., Howes, R., Wang, M., and Ferrer, C.~C.
\newblock The deepfake detection challenge (dfdc) dataset.
\newblock \emph{arXiv preprint arXiv:2006.07397}, 2020.

\bibitem[Durall et~al.(2019)Durall, Keuper, Pfreundt, and Keuper]{durall2019unmasking}
Durall, R., Keuper, M., Pfreundt, F.-J., and Keuper, J.
\newblock Unmasking deepfakes with simple features.
\newblock \emph{arXiv preprint arXiv:1911.00686}, 2019.

\bibitem[Durall et~al.(2020)Durall, Keuper, and Keuper]{durall2020watch}
Durall, R., Keuper, M., and Keuper, J.
\newblock Watch your up-convolution: Cnn based generative deep neural networks are failing to reproduce spectral distributions.
\newblock In \emph{Proceedings of the IEEE/CVF conference on computer vision and pattern recognition}, pp.\  7890--7899, 2020.

\bibitem[Fernandez et~al.(2022)Fernandez, Sablayrolles, Furon, J{\'e}gou, and Douze]{fernandez2022watermarking}
Fernandez, P., Sablayrolles, A., Furon, T., J{\'e}gou, H., and Douze, M.
\newblock Watermarking images in self-supervised latent spaces.
\newblock In \emph{ICASSP 2022-2022 IEEE International Conference on Acoustics, Speech and Signal Processing (ICASSP)}, pp.\  3054--3058. IEEE, 2022.

\bibitem[Fernandez et~al.(2023)Fernandez, Couairon, J{\'e}gou, Douze, and Furon]{fernandez2023stable}
Fernandez, P., Couairon, G., J{\'e}gou, H., Douze, M., and Furon, T.
\newblock The stable signature: Rooting watermarks in latent diffusion models.
\newblock In \emph{Proceedings of the IEEE/CVF International Conference on Computer Vision}, pp.\  22466--22477, 2023.

\bibitem[Flynn et~al.(2021)Flynn, Clough, and Cooke]{flynn2021disrupting}
Flynn, A., Clough, J., and Cooke, T.
\newblock Disrupting and preventing deepfake abuse: Exploring criminal law responses to ai-facilitated abuse.
\newblock \emph{The palgrave handbook of gendered violence and technology}, pp.\  583--603, 2021.

\bibitem[Flynn et~al.(2022)Flynn, Powell, Scott, and Cama]{flynn2022deepfakes}
Flynn, A., Powell, A., Scott, A.~J., and Cama, E.
\newblock Deepfakes and digitally altered imagery abuse: A cross-country exploration of an emerging form of image-based sexual abuse.
\newblock \emph{The British Journal of Criminology}, 62\penalty0 (6):\penalty0 1341--1358, 2022.

\bibitem[Francke \& Bennett(2019)Francke and Bennett]{francke2019potential}
Francke, E. and Bennett, A.
\newblock The potential influence of artificial intelligence on plagiarism: A higher education perspective.
\newblock In \emph{European Conference on the Impact of Artificial Intelligence and Robotics (ECIAIR 2019)}, pp.\  131--140, 2019.

\bibitem[Frank et~al.(2020)Frank, Eisenhofer, Sch{\"o}nherr, Fischer, Kolossa, and Holz]{frank2020leveraging}
Frank, J., Eisenhofer, T., Sch{\"o}nherr, L., Fischer, A., Kolossa, D., and Holz, T.
\newblock Leveraging frequency analysis for deep fake image recognition.
\newblock In \emph{International conference on machine learning}, pp.\  3247--3258. PMLR, 2020.

\bibitem[Goodfellow et~al.(2014)Goodfellow, Pouget-Abadie, Mirza, Xu, Warde-Farley, Ozair, Courville, and Bengio]{Goodfellow2014GenerativeAN}
Goodfellow, I.~J., Pouget-Abadie, J., Mirza, M., Xu, B., Warde-Farley, D., Ozair, S., Courville, A.~C., and Bengio, Y.
\newblock Generative adversarial nets.
\newblock In \emph{Advances in neural information processing systems}, 2014.

\bibitem[Grubbs(1950)]{grubbs1950sample}
Grubbs, F.~E.
\newblock Sample criteria for testing outlying observations.
\newblock \emph{The Annals of Mathematical Statistics}, pp.\  27--58, 1950.

\bibitem[Gu(2024)]{gu2024responsible}
Gu, J.
\newblock Responsible generative ai: What to generate and what not.
\newblock \emph{arXiv preprint arXiv:2404.05783}, 2024.

\bibitem[Gu et~al.(2022)Gu, Chen, Bao, Wen, Zhang, Chen, Yuan, and Guo]{gu2022vector}
Gu, S., Chen, D., Bao, J., Wen, F., Zhang, B., Chen, D., Yuan, L., and Guo, B.
\newblock Vector quantized diffusion model for text-to-image synthesis.
\newblock In \emph{Proceedings of the IEEE/CVF Conference on Computer Vision and Pattern Recognition}, pp.\  10696--10706, 2022.

\bibitem[Guo et~al.(2024)Guo, Li, Wang, Xia, Huang, Liu, and Li]{guo2024domain}
Guo, J., Li, Y., Wang, L., Xia, S.-T., Huang, H., Liu, C., and Li, B.
\newblock Domain watermark: Effective and harmless dataset copyright protection is closed at hand.
\newblock \emph{Advances in Neural Information Processing Systems}, 36, 2024.

\bibitem[Huang et~al.(2023)Huang, Li, Cai, Wang, Guo, Fang, Chen, and Wang]{huang2023can}
Huang, Z., Li, B., Cai, Y., Wang, R., Guo, S., Fang, L., Chen, J., and Wang, L.
\newblock What can discriminator do? towards box-free ownership verification of generative adversarial networks.
\newblock In \emph{Proceedings of the IEEE/CVF international conference on computer vision}, pp.\  5009--5019, 2023.

\bibitem[Inc.()]{prompthero}
Inc., P.
\newblock {PromptHero}.
\newblock \url{https://prompthero.com/}.

\bibitem[Jahanian et~al.(2019)Jahanian, Chai, and Isola]{jahanian2019steerability}
Jahanian, A., Chai, L., and Isola, P.
\newblock On the" steerability" of generative adversarial networks.
\newblock \emph{arXiv preprint arXiv:1907.07171}, 2019.

\bibitem[Jeong et~al.(2022)Jeong, Kim, Ro, and Choi]{jeong2022frepgan}
Jeong, Y., Kim, D., Ro, Y., and Choi, J.
\newblock Frepgan: robust deepfake detection using frequency-level perturbations.
\newblock In \emph{Proceedings of the AAAI Conference on Artificial Intelligence}, volume~36, pp.\  1060--1068, 2022.

\bibitem[Jovanovi{\'c} et~al.(2024)Jovanovi{\'c}, Staab, and Vechev]{jovanovic2024watermark}
Jovanovi{\'c}, N., Staab, R., and Vechev, M.
\newblock Watermark stealing in large language models.
\newblock \emph{arXiv preprint arXiv:2402.19361}, 2024.

\bibitem[Karras et~al.(2020)Karras, Laine, Aittala, Hellsten, Lehtinen, and Aila]{karras2020analyzing}
Karras, T., Laine, S., Aittala, M., Hellsten, J., Lehtinen, J., and Aila, T.
\newblock Analyzing and improving the image quality of stylegan.
\newblock In \emph{Proceedings of the IEEE/CVF conference on computer vision and pattern recognition}, pp.\  8110--8119, 2020.

\bibitem[Kietzmann et~al.(2020)Kietzmann, Lee, McCarthy, and Kietzmann]{kietzmann2020deepfakes}
Kietzmann, J., Lee, L.~W., McCarthy, I.~P., and Kietzmann, T.~C.
\newblock Deepfakes: Trick or treat?
\newblock \emph{Business Horizons}, 63\penalty0 (2):\penalty0 135--146, 2020.

\bibitem[Kingma \& Welling(2013)Kingma and Welling]{kingma2013auto}
Kingma, D.~P. and Welling, M.
\newblock Auto-encoding variational bayes.
\newblock \emph{arXiv preprint arXiv:1312.6114}, 2013.

\bibitem[Kishore et~al.(2021)Kishore, Chen, Wang, Li, and Weinberger]{kishore2021fixed}
Kishore, V., Chen, X., Wang, Y., Li, B., and Weinberger, K.~Q.
\newblock Fixed neural network steganography: Train the images, not the network.
\newblock In \emph{International Conference on Learning Representations}, 2021.

\bibitem[Krizhevsky et~al.(2009)Krizhevsky, Hinton, et~al.]{krizhevsky2009learning}
Krizhevsky, A., Hinton, G., et~al.
\newblock Learning multiple layers of features from tiny images.
\newblock 2009.

\bibitem[Laszkiewicz et~al.(2023)Laszkiewicz, Ricker, Lederer, and Fischer]{laszkiewicz2023single}
Laszkiewicz, M., Ricker, J., Lederer, J., and Fischer, A.
\newblock Single-model attribution via final-layer inversion.
\newblock \emph{arXiv preprint arXiv:2306.06210}, 2023.

\bibitem[Li et~al.(2023{\natexlab{a}})Li, Shen, Torr, Tresp, and Gu]{li2023self}
Li, H., Shen, C., Torr, P., Tresp, V., and Gu, J.
\newblock Self-discovering interpretable diffusion latent directions for responsible text-to-image generation.
\newblock \emph{arXiv preprint arXiv:2311.17216}, 2023{\natexlab{a}}.

\bibitem[Li et~al.(2023{\natexlab{b}})Li, Zhu, Yang, Jiang, Wei, and Xia]{li2023black}
Li, Y., Zhu, M., Yang, X., Jiang, Y., Wei, T., and Xia, S.-T.
\newblock Black-box dataset ownership verification via backdoor watermarking.
\newblock \emph{IEEE Transactions on Information Forensics and Security}, 2023{\natexlab{b}}.

\bibitem[Liu et~al.(2024)Liu, Khakzar, Gu, Chen, Torr, and Pizzati]{liu2024latent}
Liu, R., Khakzar, A., Gu, J., Chen, Q., Torr, P., and Pizzati, F.
\newblock Latent guard: a safety framework for text-to-image generation.
\newblock \emph{arXiv preprint arXiv:2404.08031}, 2024.

\bibitem[Liu et~al.(2020)Liu, Qi, and Torr]{liu2020global}
Liu, Z., Qi, X., and Torr, P.~H.
\newblock Global texture enhancement for fake face detection in the wild.
\newblock In \emph{Proceedings of the IEEE/CVF conference on computer vision and pattern recognition}, pp.\  8060--8069, 2020.

\bibitem[Lu et~al.(2022)Lu, Zhou, Bao, Chen, Li, and Zhu]{lu2022dpm}
Lu, C., Zhou, Y., Bao, F., Chen, J., Li, C., and Zhu, J.
\newblock Dpm-solver: A fast ode solver for diffusion probabilistic model sampling in around 10 steps.
\newblock \emph{Advances in Neural Information Processing Systems}, 35:\penalty0 5775--5787, 2022.

\bibitem[Luo et~al.(2009)Luo, Chen, Chen, Zeng, and Xiong]{luo2009reversible}
Luo, L., Chen, Z., Chen, M., Zeng, X., and Xiong, Z.
\newblock Reversible image watermarking using interpolation technique.
\newblock \emph{IEEE Transactions on information forensics and security}, 5\penalty0 (1):\penalty0 187--193, 2009.

\bibitem[Luo et~al.(2023)Luo, Tan, Huang, Li, and Zhao]{luo2023latent}
Luo, S., Tan, Y., Huang, L., Li, J., and Zhao, H.
\newblock Latent consistency models: Synthesizing high-resolution images with few-step inference.
\newblock \emph{arXiv preprint arXiv:2310.04378}, 2023.

\bibitem[Mirsky \& Lee(2021)Mirsky and Lee]{mirsky2021creation}
Mirsky, Y. and Lee, W.
\newblock The creation and detection of deepfakes: A survey.
\newblock \emph{ACM Computing Surveys (CSUR)}, 54\penalty0 (1):\penalty0 1--41, 2021.

\bibitem[Mokady et~al.(2023)Mokady, Hertz, Aberman, Pritch, and Cohen-Or]{mokady2023null}
Mokady, R., Hertz, A., Aberman, K., Pritch, Y., and Cohen-Or, D.
\newblock Null-text inversion for editing real images using guided diffusion models.
\newblock In \emph{Proceedings of the IEEE/CVF Conference on Computer Vision and Pattern Recognition}, pp.\  6038--6047, 2023.

\bibitem[Pan et~al.(2024)Pan, Wang, Dong, Sehwag, Lyu, and Lin]{pan2024finding}
Pan, M., Wang, Z., Dong, X., Sehwag, V., Lyu, L., and Lin, X.
\newblock Finding needles in a haystack: A black-box approach to invisible watermark detection.
\newblock \emph{arXiv preprint arXiv:2403.15955}, 2024.

\bibitem[Parmar et~al.(2023)Parmar, Kumar~Singh, Zhang, Li, Lu, and Zhu]{parmar2023zero}
Parmar, G., Kumar~Singh, K., Zhang, R., Li, Y., Lu, J., and Zhu, J.-Y.
\newblock Zero-shot image-to-image translation.
\newblock In \emph{ACM SIGGRAPH 2023 Conference Proceedings}, pp.\  1--11, 2023.

\bibitem[Partadiredja et~al.(2020)Partadiredja, Serrano, and Ljubenkov]{partadiredja2020ai}
Partadiredja, R.~A., Serrano, C.~E., and Ljubenkov, D.
\newblock Ai or human: the socio-ethical implications of ai-generated media content.
\newblock In \emph{2020 13th CMI Conference on Cybersecurity and Privacy (CMI)-Digital Transformation-Potentials and Challenges (51275)}, pp.\  1--6. IEEE, 2020.

\bibitem[Pereira \& Pun(2000)Pereira and Pun]{pereira2000robust}
Pereira, S. and Pun, T.
\newblock Robust template matching for affine resistant image watermarks.
\newblock \emph{IEEE transactions on image Processing}, 9\penalty0 (6):\penalty0 1123--1129, 2000.

\bibitem[Razzhigaev et~al.(2023)Razzhigaev, Shakhmatov, Maltseva, Arkhipkin, Pavlov, Ryabov, Kuts, Panchenko, Kuznetsov, and Dimitrov]{razzhigaev2023kandinsky}
Razzhigaev, A., Shakhmatov, A., Maltseva, A., Arkhipkin, V., Pavlov, I., Ryabov, I., Kuts, A., Panchenko, A., Kuznetsov, A., and Dimitrov, D.
\newblock Kandinsky: an improved text-to-image synthesis with image prior and latent diffusion.
\newblock \emph{arXiv preprint arXiv:2310.03502}, 2023.

\bibitem[Rombach et~al.(2022)Rombach, Blattmann, Lorenz, Esser, and Ommer]{rombach2022high}
Rombach, R., Blattmann, A., Lorenz, D., Esser, P., and Ommer, B.
\newblock High-resolution image synthesis with latent diffusion models.
\newblock In \emph{Proceedings of the IEEE/CVF Conference on Computer Vision and Pattern Recognition}, pp.\  10684--10695, 2022.

\bibitem[Russakovsky et~al.(2015)Russakovsky, Deng, Su, Krause, Satheesh, Ma, Huang, Karpathy, Khosla, Bernstein, et~al.]{russakovsky2015imagenet}
Russakovsky, O., Deng, J., Su, H., Krause, J., Satheesh, S., Ma, S., Huang, Z., Karpathy, A., Khosla, A., Bernstein, M., et~al.
\newblock Imagenet large scale visual recognition challenge.
\newblock \emph{International journal of computer vision}, 115\penalty0 (3):\penalty0 211--252, 2015.

\bibitem[Schramowski et~al.(2023)Schramowski, Brack, Deiseroth, and Kersting]{schramowski2023safe}
Schramowski, P., Brack, M., Deiseroth, B., and Kersting, K.
\newblock Safe latent diffusion: Mitigating inappropriate degeneration in diffusion models.
\newblock In \emph{Proceedings of the IEEE/CVF Conference on Computer Vision and Pattern Recognition}, pp.\  22522--22531, 2023.

\bibitem[Schuhmann et~al.(2022)Schuhmann, Beaumont, Vencu, Gordon, Wightman, Cherti, Coombes, Katta, Mullis, Wortsman, et~al.]{schuhmann2022laion}
Schuhmann, C., Beaumont, R., Vencu, R., Gordon, C., Wightman, R., Cherti, M., Coombes, T., Katta, A., Mullis, C., Wortsman, M., et~al.
\newblock Laion-5b: An open large-scale dataset for training next generation image-text models.
\newblock \emph{Advances in Neural Information Processing Systems}, 35:\penalty0 25278--25294, 2022.

\bibitem[Sha et~al.(2022)Sha, Li, Yu, and Zhang]{sha2022fake}
Sha, Z., Li, Z., Yu, N., and Zhang, Y.
\newblock De-fake: Detection and attribution of fake images generated by text-to-image diffusion models.
\newblock \emph{arXiv preprint arXiv:2210.06998}, 2022.

\bibitem[Shao et~al.(2024)Shao, Li, Yao, He, Qin, and Ren]{shao2024explanation}
Shao, S., Li, Y., Yao, H., He, Y., Qin, Z., and Ren, K.
\newblock Explanation as a watermark: Towards harmless and multi-bit model ownership verification via watermarking feature attribution.
\newblock \emph{arXiv preprint arXiv:2405.04825}, 2024.

\bibitem[Song et~al.(2020)Song, Meng, and Ermon]{song2020denoising}
Song, J., Meng, C., and Ermon, S.
\newblock Denoising diffusion implicit models.
\newblock \emph{arXiv preprint arXiv:2010.02502}, 2020.

\bibitem[Swanson et~al.(1996)Swanson, Zhu, and Tewfik]{swanson1996transparent}
Swanson, M.~D., Zhu, B., and Tewfik, A.~H.
\newblock Transparent robust image watermarking.
\newblock In \emph{Proceedings of 3rd IEEE International Conference on Image Processing}, volume~3, pp.\  211--214. IEEE, 1996.

\bibitem[Tancik et~al.(2020)Tancik, Mildenhall, and Ng]{tancik2020stegastamp}
Tancik, M., Mildenhall, B., and Ng, R.
\newblock Stegastamp: Invisible hyperlinks in physical photographs.
\newblock In \emph{Proceedings of the IEEE/CVF conference on computer vision and pattern recognition}, pp.\  2117--2126, 2020.

\bibitem[Tao et~al.(2022)Tao, Wang, Cheng, Ma, An, Liu, Shen, Zhang, Mao, and Zhang]{tao2022backdoor}
Tao, G., Wang, Z., Cheng, S., Ma, S., An, S., Liu, Y., Shen, G., Zhang, Z., Mao, Y., and Zhang, X.
\newblock Backdoor vulnerabilities in normally trained deep learning models.
\newblock \emph{arXiv preprint arXiv:2211.15929}, 2022.

\bibitem[Van Den~Oord et~al.(2017)Van Den~Oord, Vinyals, et~al.]{van2017neural}
Van Den~Oord, A., Vinyals, O., et~al.
\newblock Neural discrete representation learning.
\newblock \emph{Advances in neural information processing systems}, 30, 2017.

\bibitem[Wang et~al.(2021)Wang, Lin, Zhao, and Zhu]{wang2021watermark}
Wang, R., Lin, C., Zhao, Q., and Zhu, F.
\newblock Watermark faker: towards forgery of digital image watermarking.
\newblock In \emph{2021 IEEE International Conference on Multimedia and Expo (ICME)}, pp.\  1--6. IEEE, 2021.

\bibitem[Wang et~al.(2023{\natexlab{a}})Wang, Ren, Li, She, Zhang, Fang, Chen, and Wang]{wang2023free}
Wang, R., Ren, J., Li, B., She, T., Zhang, W., Fang, L., Chen, J., and Wang, L.
\newblock Free fine-tuning: A plug-and-play watermarking scheme for deep neural networks.
\newblock In \emph{Proceedings of the 31st ACM International Conference on Multimedia}, pp.\  8463--8474, 2023{\natexlab{a}}.

\bibitem[Wang et~al.(2020)Wang, Wang, Zhang, Owens, and Efros]{wang2020cnn}
Wang, S.-Y., Wang, O., Zhang, R., Owens, A., and Efros, A.~A.
\newblock Cnn-generated images are surprisingly easy to spot... for now.
\newblock In \emph{Proceedings of the IEEE/CVF conference on computer vision and pattern recognition}, pp.\  8695--8704, 2020.

\bibitem[Wang et~al.(2004)Wang, Bovik, Sheikh, and Simoncelli]{wang2004image}
Wang, Z., Bovik, A.~C., Sheikh, H.~R., and Simoncelli, E.~P.
\newblock Image quality assessment: from error visibility to structural similarity.
\newblock \emph{IEEE transactions on image processing}, 13\penalty0 (4):\penalty0 600--612, 2004.

\bibitem[Wang et~al.(2022{\natexlab{a}})Wang, Mei, Zhai, and Ma]{wang2022unicorn}
Wang, Z., Mei, K., Zhai, J., and Ma, S.
\newblock Unicorn: A unified backdoor trigger inversion framework.
\newblock In \emph{The Eleventh International Conference on Learning Representations}, 2022{\natexlab{a}}.

\bibitem[Wang et~al.(2022{\natexlab{b}})Wang, Zhai, and Ma]{wang2022bppattack}
Wang, Z., Zhai, J., and Ma, S.
\newblock Bppattack: Stealthy and efficient trojan attacks against deep neural networks via image quantization and contrastive adversarial learning.
\newblock In \emph{Proceedings of the IEEE/CVF Conference on Computer Vision and Pattern Recognition}, pp.\  15074--15084, 2022{\natexlab{b}}.

\bibitem[Wang et~al.(2023{\natexlab{b}})Wang, Bao, Zhou, Wang, Hu, Chen, and Li]{wang2023dire}
Wang, Z., Bao, J., Zhou, W., Wang, W., Hu, H., Chen, H., and Li, H.
\newblock Dire for diffusion-generated image detection.
\newblock \emph{arXiv preprint arXiv:2303.09295}, 2023{\natexlab{b}}.

\bibitem[Wang et~al.(2023{\natexlab{c}})Wang, Chen, Lyu, Metaxas, and Ma]{wang2023diagnosis}
Wang, Z., Chen, C., Lyu, L., Metaxas, D.~N., and Ma, S.
\newblock Diagnosis: Detecting unauthorized data usages in text-to-image diffusion models.
\newblock In \emph{The Twelfth International Conference on Learning Representations}, 2023{\natexlab{c}}.

\bibitem[Wang et~al.(2023{\natexlab{d}})Wang, Chen, Zeng, Lyu, and Ma]{wang2023origin}
Wang, Z., Chen, C., Zeng, Y., Lyu, L., and Ma, S.
\newblock Where did i come from? origin attribution of ai-generated images.
\newblock \emph{Advances in neural information processing systems}, 2023{\natexlab{d}}.

\bibitem[Wen et~al.(2023{\natexlab{a}})Wen, Kirchenbauer, Geiping, and Goldstein]{wen2023tree}
Wen, Y., Kirchenbauer, J., Geiping, J., and Goldstein, T.
\newblock Tree-ring watermarks: Fingerprints for diffusion images that are invisible and robust.
\newblock \emph{arXiv preprint arXiv:2305.20030}, 2023{\natexlab{a}}.

\bibitem[Wen et~al.(2023{\natexlab{b}})Wen, Liu, Chen, and Lyu]{wen2023detecting}
Wen, Y., Liu, Y., Chen, C., and Lyu, L.
\newblock Detecting, explaining, and mitigating memorization in diffusion models.
\newblock In \emph{The Twelfth International Conference on Learning Representations}, 2023{\natexlab{b}}.

\bibitem[Whittaker et~al.(2020)Whittaker, Kietzmann, Kietzmann, and Dabirian]{whittaker2020all}
Whittaker, L., Kietzmann, T.~C., Kietzmann, J., and Dabirian, A.
\newblock “all around me are synthetic faces”: the mad world of ai-generated media.
\newblock \emph{IT Professional}, 22\penalty0 (5):\penalty0 90--99, 2020.

\bibitem[Wu et~al.(2024)Wu, Ma, Wang, Zhang, Liang, Li, Lin, Fang, and Wang]{wu2024traceevader}
Wu, M., Ma, J., Wang, R., Zhang, S., Liang, Z., Li, B., Lin, C., Fang, L., and Wang, L.
\newblock Traceevader: Making deepfakes more untraceable via evading the forgery model attribution.
\newblock In \emph{Proceedings of the AAAI Conference on Artificial Intelligence}, volume~38, pp.\  19965--19973, 2024.

\bibitem[Yu et~al.(2021{\natexlab{a}})Yu, Li, Koh, Zhang, Pang, Qin, Ku, Xu, Baldridge, and Wu]{yu2021vector}
Yu, J., Li, X., Koh, J.~Y., Zhang, H., Pang, R., Qin, J., Ku, A., Xu, Y., Baldridge, J., and Wu, Y.
\newblock Vector-quantized image modeling with improved vqgan.
\newblock \emph{arXiv preprint arXiv:2110.04627}, 2021{\natexlab{a}}.

\bibitem[Yu et~al.(2022{\natexlab{a}})Yu, Xu, Koh, Luong, Baid, Wang, Vasudevan, Ku, Yang, Ayan, et~al.]{yu2022scaling}
Yu, J., Xu, Y., Koh, J.~Y., Luong, T., Baid, G., Wang, Z., Vasudevan, V., Ku, A., Yang, Y., Ayan, B.~K., et~al.
\newblock Scaling autoregressive models for content-rich text-to-image generation.
\newblock \emph{arXiv preprint arXiv:2206.10789}, 2\penalty0 (3):\penalty0 5, 2022{\natexlab{a}}.

\bibitem[Yu et~al.(2019)Yu, Davis, and Fritz]{yu2019attributing}
Yu, N., Davis, L.~S., and Fritz, M.
\newblock Attributing fake images to gans: Learning and analyzing gan fingerprints.
\newblock In \emph{Proceedings of the IEEE/CVF international conference on computer vision}, pp.\  7556--7566, 2019.

\bibitem[Yu et~al.(2021{\natexlab{b}})Yu, Skripniuk, Abdelnabi, and Fritz]{yu2021artificial}
Yu, N., Skripniuk, V., Abdelnabi, S., and Fritz, M.
\newblock Artificial fingerprinting for generative models: Rooting deepfake attribution in training data.
\newblock In \emph{Proceedings of the IEEE/CVF International conference on computer vision}, pp.\  14448--14457, 2021{\natexlab{b}}.

\bibitem[Yu et~al.(2022{\natexlab{b}})Yu, Skripniuk, Chen, Davis, and Fritz]{yu2022responsible}
Yu, N., Skripniuk, V., Chen, D., Davis, L.~S., and Fritz, M.
\newblock Responsible disclosure of generative models using scalable fingerprinting.
\newblock In \emph{International Conference on Learning Representations}, 2022{\natexlab{b}}.

\bibitem[Zhang et~al.(2020)Zhang, Zhou, Shumailov, and Papernot]{zhang2020attribution}
Zhang, B., Zhou, J.~P., Shumailov, I., and Papernot, N.
\newblock On attribution of deepfakes.
\newblock \emph{arXiv preprint arXiv:2008.09194}, 2020.

\bibitem[Zhao et~al.(2021)Zhao, Zhou, Chen, Wei, Zhang, and Yu]{zhao2021multi}
Zhao, H., Zhou, W., Chen, D., Wei, T., Zhang, W., and Yu, N.
\newblock Multi-attentional deepfake detection.
\newblock In \emph{Proceedings of the IEEE/CVF conference on computer vision and pattern recognition}, pp.\  2185--2194, 2021.

\bibitem[Zheng et~al.(2022)Zheng, Vuong, Cai, and Phung]{zheng2022movq}
Zheng, C., Vuong, T.-L., Cai, J., and Phung, D.
\newblock Movq: Modulating quantized vectors for high-fidelity image generation.
\newblock \emph{Advances in Neural Information Processing Systems}, 35:\penalty0 23412--23425, 2022.

\bibitem[Zhu et~al.(2018)Zhu, Kaplan, Johnson, and Fei-Fei]{zhu2018hidden}
Zhu, J., Kaplan, R., Johnson, J., and Fei-Fei, L.
\newblock Hidden: Hiding data with deep networks.
\newblock In \emph{Proceedings of the European conference on computer vision (ECCV)}, pp.\  657--672, 2018.

\bibitem[Zhu et~al.(2016)Zhu, Kr{\"a}henb{\"u}hl, Shechtman, and Efros]{zhu2016generative}
Zhu, J.-Y., Kr{\"a}henb{\"u}hl, P., Shechtman, E., and Efros, A.~A.
\newblock Generative visual manipulation on the natural image manifold.
\newblock In \emph{Computer Vision--ECCV 2016: 14th European Conference, Amsterdam, The Netherlands, October 11-14, 2016, Proceedings, Part V 14}, pp.\  597--613. Springer, 2016.

\bibitem[Zou et~al.(2023)Zou, Wang, Kolter, and Fredrikson]{zou2023universal}
Zou, A., Wang, Z., Kolter, J.~Z., and Fredrikson, M.
\newblock Universal and transferable adversarial attacks on aligned language models.
\newblock \emph{arXiv preprint arXiv:2307.15043}, 2023.

\end{thebibliography}

\clearpage

\section{Appendix}

\subsection{More Details of the Used Models}
\label{sec:appendix_model_details}

In this section, we provide more details about the model used in the experiments.

\noindent
\textbf{Stable Diffusion v1-4\footnote{https://huggingface.co/CompVis/stable-diffusion-v1-4}.} This model is initialized with the weights of the Stable-Diffusion-v1-2 model\footnote{https://huggingface.co/CompVis/stable-diffusion-v1-2} and then fine-tuned on 225k steps at resolution 512x512 on "laion-aesthetics v2 5+". The architecture of the auto-encoder used in this model is the VAE. This model is with creativeml-openrail-m License.

\noindent
\textbf{Stable Diffusion v1-5\footnote{https://huggingface.co/runwayml/stable-diffusion-v1-5}.} This model is initialized with the weights of the Stable-Diffusion-v1-2 model and subsequently fine-tuned on 595k steps on "laion-aesthetics v2 5+" with 512x512 resolution. The architecture of the auto-encoder used in this model is the VAE.
This model is with creativeml-openrail-m License.

\noindent
\textbf{Stable Diffusion v2-base\footnote{https://huggingface.co/stabilityai/stable-diffusion-2-base}.} This model model is trained from scratch 550k steps at resolution 256x256 on a subset of LAION-5B filtered for explicit pornographic material, using the LAION-NSFW classifier with punsafe=0.1 and an aesthetic score $\geq$ 4.5. It is further trained for 850k steps at resolution 512x512 on the same dataset on images with resolution $\geq$ 512x512. The architecture of the auto-encoder used in this model is the VAE.
This model is with openrail++ License.

\noindent
\textbf{Stable Diffusion v2-1\footnote{https://huggingface.co/stabilityai/stable-diffusion-2-1}.} This model is fine-tuned from Stable Diffusion 2 model\footnote{https://huggingface.co/stabilityai/stable-diffusion-2} with an additional 55k steps on a subset of LAION-5B filtered for explicit pornographic material (with punsafe=0.1), and then fine-tuned for another 155k extra steps with punsafe=0.98. The architecture of the auto-encoder used in this model is the VAE.
This model is with openrail++ License.

\noindent
\textbf{Stable Diffusion XL-1.0-base\footnote{https://huggingface.co/stabilityai/stable-diffusion-xl-base-1.0}.}
This model is first trained from scratch on an internal dataset constructed by Stability-AI for 600 000 optimization steps at a resolution of 256 × 256 pixels and a batch-size of 2048. Then, it is trained on 512 × 512 pixel images for
another 200 000 optimization steps. Finally, the trainers utilize multi-aspect training in combination
with an offset-noise level of 0.05 to train the model on different aspect ratios of around 1024 × 1024 pixel area. The developers train the same auto-encoder architecture used for the original Stable Diffusion at
a larger batch-size (256 vs 9) and additionally
track the weights with an exponential moving
average. This model is with openrail++ License.

\noindent
\textbf{Kandinsky 2.1\footnote{https://huggingface.co/ai-forever/Kandinsky\_2.1}.}
This model utilizes CLIP and diffusion image prior (mapping) between latent spaces of CLIP modalities to increase the generation performance.
For diffusion mapping of latent spaces, it uses the transformer architecture with num\_layers=20, num\_heads=32 and hidden\_size=2048. It also uses the custom implementation of MoVQGAN~\cite{zheng2022movq} with minor modifications as the autoencoder~\cite{razzhigaev2023kandinsky}. The autoencoder is trained on the LAION HighRes dataset~\cite{schuhmann2022laion}.
This model is with apache-2.0 License.

\subsection{Critical Value of the t-Distribution}
\label{sec:appendix_critical}

In this section, we introduce the detailed process for calculating the 
critical value of the t-distribution, which is used in \autoref{eq:hypo} for obtaining the detection threshold. The probability density function for the t-distribution is written in \autoref{eq:pdf}, where \(\nu\)  is the number of degrees of freedom and 
\(\Gamma\)  is the gamma function.

\begin{equation}
\label{eq:pdf}
f(a)=\frac{\Gamma\left(\frac{\nu+1}{2}\right)}{\sqrt{\nu \pi} \Gamma\left(\frac{\nu}{2}\right)}\left(1+\frac{a^2}{\nu}\right)^{-(\nu+1) / 2}
\end{equation}

\noindent
Then, the corresponding cumulative distribution function is written in \autoref{eq:cdf}, where \(\beta\) denotes the incomplete beta function.

\begin{equation}
\label{eq:cdf}
\mathbb{P}(a<t^{\prime})=\int_{-\infty}^{t^{\prime}} f(u) d u=1-\frac{1}{2} \beta\left(\frac{\nu}{{t^{\prime}}^2 + \nu},\frac{\nu}{2}, \frac{1}{2}\right)
\end{equation}

\noindent
Thus, given a confidence level \(\alpha\) and the number of degrees of freedom \(\nu\), we have can use \autoref{eq:tav} to calculate the value of the critical value \(t_{\alpha, \nu}\).

\begin{equation}
\label{eq:tav}
\mathbb{P}(a<t_{\alpha, \nu}) = 1 - \alpha
\end{equation}

\subsection{More Results about the Effectiveness}
\label{sec:appendix_auroc}
In this section, we report the AUROC of our method to fully understand its performance. The results are shown in \autoref{tab:auroc}.
The results demonstrate that our method achieves high AUROC in all settings, indicating its high performance and non-sensitivity to different threshold selection methods.

\begin{table}[]
\centering
\scriptsize
\setlength\tabcolsep{3pt}
\caption{AUROC for distinguishing belonging images and images generated by other models. Here, Model \(\mathcal{M}_1\) is the inspected model, Model \(\mathcal{M}_2\) is the other model.}\label{tab:auroc}
\begin{tabular}{@{}ccc@{}}
\toprule
Model $\mathcal{M}_1$   & Model $\mathcal{M}_2$       & AUROC \\ \midrule
SD v1-4    & SD v2-base     & 0.98  \\
SD v1-4    & SD v2-1        & 0.99  \\
SD v1-4    & SD XL-1.0-base & 0.96  \\
SD v1-4    & Kandinsky      & 0.98  \\
SD v1-5    & SD v2-base     & 0.98  \\
SD v1-5    & SD v2-1        & 0.99  \\
SD v1-5    & SD XL-1.0-base & 0.97  \\
SD v1-5    & Kandinsky      & 0.98  \\
SD v2-base & SD v1-4        & 0.99  \\
SD v2-base & SD v1-5        & 0.99  \\
SD v2-base & SD XL-1.0-base & 0.98  \\
SD v2-base & Kandinsky      & 0.99  \\
Kandinsky  & SD v1-4        & 0.99  \\
Kandinsky  & SD v1-5        & 0.99  \\
Kandinsky  & SD v2-base     & 0.98  \\
Kandinsky  & SD v2-1        & 0.98  \\
Kandinsky  & SD XL-1.0-base & 0.97  \\ \bottomrule
\end{tabular}
\end{table}

\subsection{Stopping Strategy}
\label{sec:stopping}

In our implementation, we set 100 as the maximum step. As shown in \autoref{fig:compare_random_encoder_losslist_encoder3}, the reconstruction loss is converged at step 100. We also evaluated the detection accuracy under an adaptive stop strategy. In detail, the reverse-engineering process stops if the reconstruction loss does not decrease in five successive steps. The results for the fixed maximum step and the adaptive stop strategy are shown in \autoref{tab:stopping}. The setting here is distinguishing the images generated by the inspected model SD v2-base and another model SD v1-5.

\begin{table}[]
\centering
\scriptsize
\setlength\tabcolsep{3pt}
\caption{Results on different stopping strategies.}\label{tab:stopping}
\begin{tabular}{@{}cc@{}}
\toprule
Strategy           & Acc    \\ \midrule
Fixed Maximum Step & 98.0\% \\
Adaptive Stop      & 97.7\% \\ \bottomrule
\end{tabular}
\end{table}

\subsection{Robustness}
\label{sec:robustness}

 We conducted the experiments for evaluating the robustness of our proposed method against a wide range of post-processing techniques employed in the referenced papers. These techniques include adjusting saturation, adjusting contrast, adding Gaussian noise, applying JPEG compression, adjusting brightness, applying Gaussian blur, and Cropping. The inspected model used here is SD v2-base. The setting here is distinguishing the images generated by the inspected model and SD v1-5 model. The other experimental settings are identical to those used in \autoref{tab:different_latent_archs}. For each experiment, we report the detection accuracy and the average values of the Structural Similarity Index (SSIM), Peak Signal-to-Noise Ratio (PSNR), L1 distance, and L2 distance between the original samples and the post-processed samples. The L1 and L2 distances are calculated using pixel values ranging from 0 to 1. 
 The results are shown in \autoref{tab:robustness}, demonstrating that our proposed method remains effective when the quality of the post-processed images is satisfactory. For instance, the detection accuracy of our method consistently exceeds 90\% when the Structural Similarity Index (SSIM) value between the original and post-processed samples is around 0.9. It is important to note that an SSIM value lower than 0.9 is considered a significant alteration to the images and indicates an unsatisfactory level of image quality~\cite{wang2004image}. In cases of strong post-processing, an adaptive attacker may be able to evade our method, but at the cost of substantially compromising the quality of the edited image. Consequently, our method maintains its effectiveness against adaptive attacks aimed at preserving the quality of the perturbed image. 

\begin{table}[]
\centering
\scriptsize
\setlength\tabcolsep{3pt}
\caption{Robustness against post-processing augmentations.}\label{tab:robustness}
\begin{tabular}{@{}cccccc@{}}
\toprule
Augmentation                & Acc    & SSIM   & PSNR    & L1     & L2         \\ \midrule
Saturation Factor 1.25      & 97.4\% & 0.9657 & 32.7953 & 0.0180 & 0.0008     \\
Saturation Factor 1.50      & 94.0\% & 0.9276 & 27.3419 & 0.0334 & 0.0028     \\
Saturation Factor 1.75      & 93.1\% & 0.8929 & 24.3338 & 0.0471 & 0.0055     \\
Saturation Factor 2.00      & 90.6\% & 0.8611 & 22.3084 & 0.0593 & 0.0085     \\
Saturation Factor 2.25      & 88.3\% & 0.8318 & 20.8090 & 0.0704 & 0.0117     \\
Saturation Factor 2.50      & 84.8\% & 0.8045 & 19.6402 & 0.0804 & 0.0150     \\ \midrule
Contrast Factor 1.10        & 95.5\% & 0.9488 & 32.6094 & 0.0198 & 0.0006     \\
Contrast Factor 1.15        & 92.7\% & 0.9213 & 29.4854 & 0.0283 & 0.0012     \\
Contrast Factor 1.25        & 89.3\% & 0.8716 & 25.7423 & 0.0433 & 0.0028     \\
Contrast Factor 1.50        & 79.8\% & 0.7729 & 21.1741 & 0.0729 & 0.0079     \\ \midrule
Gaussian Noise Std 0.01     & 97.4\% & 0.9896 & 46.1284 & 0.0039 & 2.4399e-05 \\
Gaussian Noise Std 0.02     & 95.2\% & 0.9612 & 40.1485 & 0.0078 & 9.6707e-05 \\
Gaussian Noise Std 0.03     & 93.3\% & 0.9204 & 36.6620 & 0.0116 & 0.0002     \\
Gaussian Noise Std 0.04     & 87.8\% & 0.8731 & 34.1959 & 0.0154 & 0.0004     \\
Gaussian Noise Std 0.05     & 80.7\% & 0.8238 & 32.2881 & 0.0192 & 0.0006     \\ \midrule
JPEG Compression Quality 95 & 96.5\% & 0.9730 & 38.6492 & 0.0082 & 0.0002     \\
JPEG Compression Quality 90 & 95.2\% & 0.9576 & 36.4326 & 0.0107 & 0.0003     \\
JPEG Compression Quality 80 & 94.1\% & 0.9335 & 33.9629 & 0.0142 & 0.0005     \\
JPEG Compression Quality 70 & 90.8\% & 0.9155 & 32.5805 & 0.0165 & 0.0006     \\
JPEG Compression Quality 60 & 90.2\% & 0.9009 & 31.6258 & 0.0184 & 0.0008     \\
JPEG Compression Quality 50 & 89.3\% & 0.8885 & 30.9288 & 0.0199 & 0.0009     \\ \midrule
Brightness Factor 1.25      & 92.6\% & 0.9410 & 19.8790 & 0.0855 & 0.0108     \\
Brightness Factor 1.35      & 90.7\% & 0.9042 & 17.4920 & 0.1129 & 0.0188     \\
Brightness Factor 1.50      & 81.9\% & 0.8471 & 15.1579 & 0.1485 & 0.0322     \\ \midrule
Gaussian Blur Box Size 1    & 96.3\% & 0.8755 & 28.9077 & 0.0218 & 0.0017     \\
Gaussian Blur Box Size 2    & 88.7\% & 0.7427 & 25.0875 & 0.0346 & 0.0041     \\
Gaussian Blur Box Size 3    & 78.4\% & 0.6625 & 23.3526 & 0.0432 & 0.0062     \\ \bottomrule
\end{tabular}
\end{table}

\subsection{Comparison to~\citet{laszkiewicz2023single}}
\label{sec:compare_laszkiewicz}

The related work by~\citet{laszkiewicz2023single} also concentrates on detecting images generated by a specific model without relying on watermarks.
In this section, we conduct the experiments for comparing our method to \citet{laszkiewicz2023single}.
The Inspected model ($\mathcal{M}_1$) here is SD v2-base. We consider the distinguishing the images generated by the Inspected model ($\mathcal{M}_1$) and that generated by different other models ($\mathcal{M}_2$). Here, SD v1-1+ means a fine-tuned version of the Stable Diffusion v1-1. The comparison results are reported in \autoref{tab:compare_las}.
The results demonstrate that our method outperforms the \citet{laszkiewicz2023single} It is understandable as the approach by \citet{laszkiewicz2023single} solely reverses the final layer of the model, causing it to lack access to some of the valuable information encoded within the weights of the other layers. In addition, \citet{laszkiewicz2023single} assume the last layer of the inspected model is invertible. Therefore, it is not applicable to the models that use non-invertible activation functions (such as the commonly-used ReLU function) in the last layer. Our method does not have this assumption. 

\begin{table*}[]
\centering
\scriptsize
\setlength\tabcolsep{3pt}
\caption{Comparison to 
\citet{laszkiewicz2023single}. Here, Model \(\mathcal{M}_1\) is the inspected model, Model \(\mathcal{M}_2\) is the other model.}\label{tab:compare_las}
\begin{tabular}{@{}ccccc@{}}
\toprule
Method             & $\mathcal{M}_1$:SD v2-base; $\mathcal{M}_2$:SD v1-1 & $\mathcal{M}_1$:SD v2-base; $\mathcal{M}_2$:SD v1-1+ & $\mathcal{M}_1$:SD v2-base; $\mathcal{M}_2$:SD v1-4 & $\mathcal{M}_1$:SD v2-base; $\mathcal{M}_2$:SD v1-5 \\ \midrule
\citet{laszkiewicz2023single} & 92.7\%                    & 90.6\%                     & 92.1\%                    & 92.2\%                    \\
\sys (Ours)               & 98.2\%                    & 98.0\%                     & 98.0\%                    & 97.8\%                    \\ \bottomrule
\end{tabular}
\end{table*}

\subsection{Examples of Failure Cases}
\label{sec:example_failure}

\begin{figure}[]
	\centering
	\footnotesize
	\includegraphics[width=1\columnwidth]{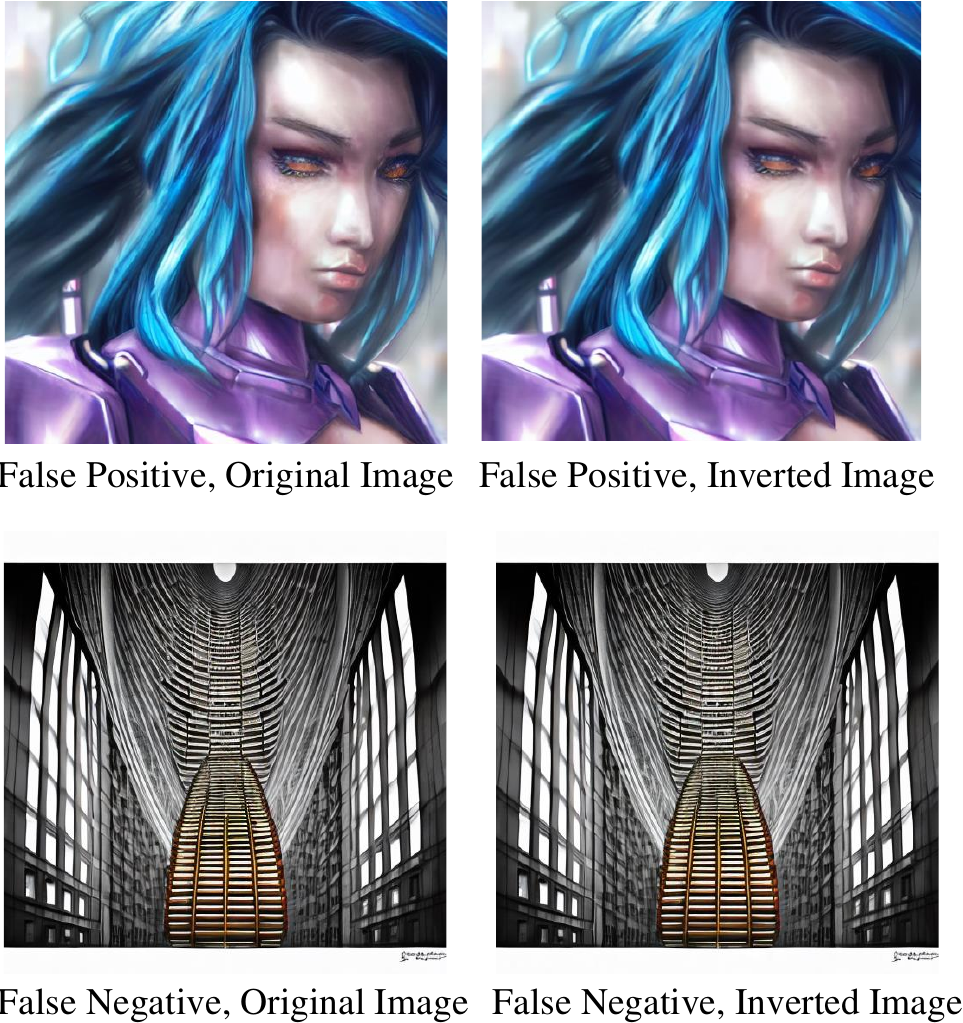}	
 \vspace{-0.3cm}
 \caption{Examples of the false positive and the false negative of our method.
 }
 \label{fig:example_failure}
 \vspace{-0.4cm}
\end{figure}

In this section, we demonstrate the examples of the failure cases of our method. The visualization results of the failure cases when distinguishing the images generated by SD-v2-base and SD-v1-5 can be found in \autoref{fig:example_failure}. We find that the potential reason for the FP and FN could be high brightness and high shape complexity in the images, respectively.

\subsection{Details of the Used Text Prompts}
\label{sec:detailed_prompts}
In \autoref{fig:compare} and \autoref{tab:different_latent_archs}, we use 50 text prompts randomly sampled from PromptHero~\cite{prompthero}. 
The detailed text prompts used can be found in
\autoref{tab:prompts1} and \autoref{tab:prompts2}.
This prompts are with MIT License. They do not contain personally identifiable information or offensive content.

\begin{table*}[]
\centering
\scriptsize
\setlength\tabcolsep{3pt}
\caption{Text prompts used in \autoref{fig:compare} and \autoref{tab:different_latent_archs} (Prompt 1-25).}
\label{tab:prompts1}
\scalebox{1}{
\begin{tabular}{@{}ll@{}}
\toprule
Prompt 1  & ``cyber punk robot, dark soul blood borne boss, face hidden, RTX technology, high resolution, light scttering"\\\toprule
Prompt 2  & \makecell[l]{``RAW photo of young woman in sun glasses sitting on beach, (closed mouth:1.2), film grain, high quality, Nikon D850, hyperrealism, photography,\\ (realistic, photo realism:1. 37), (highest quality)"}\\\toprule
Prompt 3  & \makecell[l]{``full color portrait of bosnian (bosniak) woman wearing a keffiyeh, epic character composition, by ilya kuvshinov, terry richardson, annie leibovitz,\\ sharp focus, natural lighting, subsurface scattering, f2, 35mm, film grain, award winning,  8k"}\\\toprule
Prompt 4  & \makecell[l]{``A silhouette of a woman of dark fantasy standing on the ground, in the style of dark navy and dark emerald, pigeoncore, heavily textured,\\ avian - themed, realistic figures, medieval - inspired, photorealistic painting"} \\\toprule
Prompt 5  & \makecell[l]{``stained glass art of goddess, mosaic-stained glass art, stained-glass illustration, close up, portrait, concept art, (best quality, masterpiece, ultra-detailed,\\ centered, extremely fine and aesthetically beautiful, super fine illustration), centered, epic composition, epic proportions, intricate, fractal art, zentangle,\\ hyper maximalism"} \\\toprule
Prompt 6  & \makecell[l]{``An artistic composition featuring a person with orange hair casually squatting in a park, capturing their nonchalant expression, hot pants, The focus\\ is on their unique sense of style, particularly their white panties peeking out from under their attire."} \\\toprule                             
Prompt 7  & \makecell[l]{``a woman holding a glowing ball in her hands, featured on cgsociety, fantasy art, very long flowing red hair, holding a pentagram shield, looks a bit\\ similar to amy adams, lightning mage spell icon, benevolent android necromancer, high priestess tarot card, anime goddess, portrait of celtic goddess\\ diana, featured on artstattion"} \\\toprule
Prompt 8  & ``masterpiece, girl alone, solo, incredibly absurd, hoodie, headphones, street, outdoor, rain, neon,"\\\toprule
Prompt 9  & \makecell[l]{``Halvard, druid, Spring, green, yellow, red, vibrant, wild, wildflowers masterpiece, shadows, expert, insanely detailed, 4k resolution, intricate detail,\\ art inspired by diego velazquez eugene delacroix''} \\\toprule
Prompt 10 & \makecell[l]{``(8k, RAW photo, high sensitivity, best quality, masterpiece, ultra high resolution, fidelity: 1.25), upper body, cat ears, (night), rain, walk, city lights,\\ delicate face, wet white shirt"} \\\toprule
Prompt 11 & ``masterpiece, centered, dynamic pose, 1girl, cute, calm, intelligent, red wavy hair, standing, batik swimsuit, beach background,"\\\toprule
Prompt 12 & \makecell[l]{``masterpiece, award winning, best quality, high quality, extremely detailed, cinematic shot, 1girl, adventurer, riding on a dragon,\\ fantasy theme, HD, 64K"} \\\toprule
Prompt 13 & \makecell[l]{``((masterpiece:1.4, best quality))+, (ultra detailed)+, blue hair , wolfcut, pink eyes, 1 girl,cyberpunk city,flat chest,wavy hair,mecha clothes,(robot girl),\\cool movement,silver bodysuit,colorful background,rainy days,(lightning effect),silver dragon armour,(cold face),cowboy shot"} \\\toprule        
Prompt 14 & \makecell[l]{``masterpiece, centered, concept art, wide shot, art nouveau, skyscraper, architecture, modern, sleek design, photography, raw photo, sharp focus,\\ vibrant illustrations, award winning"} \\\toprule
Prompt 15 & \makecell[l]{``masterpiece, best quality, mid shot, front view, concept art, 1girl, warrior outfit, pretty, medium blue wavy hair, walking, curious, exploring city,\\ london city street background, Fantasy theme, depth of field, global illumination, (epic composition, epic proportion), Award winning, HD, Panoramic,"} \\\toprule
Prompt 16 & \makecell[l]{``a couple of women standing next to each other holding candles, inspired by WLOP, cgsociety contest winner, ancient libu young girl, 4 k detail,\\ dressed in roman clothes, lovely detailed faces, loli, high detailed 8 k, twin souls, cgsociety, beautiful maiden"} \\\toprule
Prompt 17 & \makecell[l]{``Tigrex from monster hunter, detailed scales, detailed eyes, anatomically correct, UHD, highly detailed, raytracing, vibrant, beautiful, expressive,\\ masterpiece, oil painting"} \\\toprule
Prompt 18 & \makecell[l]{``Fashion photography of a joker, 1800s renaissance, clown makeup, editorial, insanely detailed and intricate, hyper-maximal, elegant, hyper-realistic,\\ warm lighting, photography, photorealistic, 8k"} \\\toprule        
Prompt 19 & ``octane render of cyberpunk batman by Tsutomu nihei, chrome silk with intricate ornate weaved golden filiegree, dark mysterious background --v 4 --q 2"    \\\toprule     
Prompt 20 & ``a cat with a (pirate hat:1.2) on a tropical beach, $\sim$*$\sim$Enhance$\sim$*$\sim$, in the style of Clyde Caldwell, vibrant colors"\\\toprule                      
Prompt 21 & \makecell[l]{``masterpiece, portrait, medium shot, cel shading style, centered image, ultra detailed illustration of Hatsune Miku of cool posing, inkpunk, ink lines,\\ strong outlines, bold traces, unframed, high contrast, cel-shaded, vector, 32k resolution, best quality"}\\\toprule    
Prompt 22 & ``((A bright vivid chaotic cyberpunk female, Fantastic and mysterious, full makeup, blue sky hair, (nature and magic), electronic eyes, fantasy world))\\\toprule     
Prompt 23 & \makecell[l]{``broken but unstoppable masked samurai in full battle gear, digital illustration, brutal epic composition, (expressionism style:1. 1), emotional, dramatic,\\ gloomy, 8k, high quality, unforgettable, emotional depth"}\\\toprule
Prompt 24 & \makecell[l]{``studio lighting, film, movie scene, extreme detail, 12k, masterpiece, hyperrealistic, realistic, Canon EOS R6 Mark II, a dragon made out of flowers\\ and leaves, beautiful gold flecks, colorful paint, golden eye, detailed body, detailed eye, multiple colored flowers"}\\\toprule
Prompt 25 & \makecell[l]{``Photo realistic young Farscape Chiana, kissy face, full Farscape Chiana white face paint, black shadowy eye makeup, white/gray lips, close-up shot, thin,\\ fit, Fashion Pose, DSLR, F/2. 8, Lens Flare, 5D, 16k, Super-Resolution, highly detailed, cinematic lighting"}\\
\bottomrule
\end{tabular}}
\end{table*}

\begin{table*}[]
\centering
\scriptsize
\setlength\tabcolsep{3pt}
\caption{Text prompts used in \autoref{fig:compare} and \autoref{tab:different_latent_archs} (Prompt 26-50).}
\label{tab:prompts2}
\scalebox{1}{
\begin{tabular}{@{}ll@{}}
\toprule
Prompt 26 & \makecell[l]{``A retro vintage Comic style poster, of a post apocalyptic universe, of a muscle car, extreme color scheme, action themed, driving on a desert road wasteland,\\ fleeting, chased by a giant fire breathing serpent like fantasy creature, in action pose, highly detailed digital art, jim lee"}\\\toprule
Prompt 27 & \makecell[l]{``cinematic CG 8k wallpaper, action scene from GTA V game, perfect symmetric cars bodies and elements, wheels rotating, real physics based image,\\ extremely detailed 4k digital painting (design trending on (Agnieszka Doroszewicz), Behance, Andrey Tkachenko, GTA V game, artstation, BMW X6\\ realistic design"}\\\toprule
Prompt 28 & \makecell[l]{``the Hulk in his Worldbreaker form, his power and rage reach astronomical levels, amidst a cityscape in ruins, reflecting the destruction he can unleash"}\\\toprule
Prompt 29 & \makecell[l]{``masterpiece, portrait, medium shot, cel shading style, centered image, ultra detailed illustration of Hatsune Miku of cool posing, inkpunk, ink lines,\\ strong outlines, bold traces, unframed, high contrast, cel-shaded, vector, 32k resolution, best quality"}\\\toprule
Prompt 30 & \makecell[l]{``Renaissance-style portrait of an astronaut in space, detailed starry background, reflective helmet."}\\\toprule
Prompt 31 & \makecell[l]{``A photo of a very intimidating orc on a battlefield, cinematic, melancholy, dynamic lighting, dark background"}\\\toprule
Prompt 32 & ``A dark fantasy devil predator, photographic, ultra detail, full detail, 8k best quality, realistic, 8k, micro intricate details"\\\toprule
Prompt 33 &  \makecell[l]{``Hello darkness, my old friend, I've come to talk to you again, heart-wrenching composition, digital painting, (expressionism:1. 1), (dramatic, gloomy,\\ emotionally profound:1. 1), intense and brooding dark tones, exceptionally high quality, high-resolution, leaving an indelible and haunting impression\\ on psyche, unforgettable, masterpiece"}\\\toprule
Prompt 34 & \makecell[l]{``epic, masterpiece, alien friendly standing on moon, intricated organic neural clothes, galactic black hole background, \{expansive:2\}\\ hyper realistic, octane, ultra detailed, 32k, raytracing"}\\\toprule
Prompt 35 & ``Geometrical art of autumn landscape, warm colors, a work of art, grotesque, Mysterious"\\\toprule
Prompt 36 & \makecell[l]{``a girl with face painting and a golden background is wearing makeup, absurd, creative, glamorous surreal, in the style of zbrush, black and white\\ abstraction, daz3d, porcelain, striking symmetrical patterns, close-up --ar 69:128 --s 750"}\\\toprule
Prompt 37 & \makecell[l]{``Forest, large tree, river in the middle, full blue moon, star's, night , haze, ultra-detailed, film photography, light leaks, Larry Bud Melman,\\ trending on artstation, sharp focus, studio photo, intricate details, highly detailed, by greg rutkowski"}\\\toprule
Prompt 38 & \makecell[l]{``a futuristic spacecraft winging through the sky, orange and beige color, in the style of realistic lifelike figures, ravencore, hispanicore, liquid metal,\\ greeble, high definition, manticore, photo, digital art, science fiction --v 5. 2"}\\\toprule
Prompt 39 & \makecell[l]{``a close up of a person with a sword, a character portrait by Hasegawa Settan, featured on cg society, antipodeans, reimagined by industrial light\\ and magic, sabattier effect, character"}\\\toprule
Prompt 40 & ``Dystopian New York, gritty, octane render, ultra-realistic, cinematic --ar 68:128 --s 750 --v 5. 2"\\\toprule
Prompt 41 & \makecell[l]{``ALIEN SPACECRAFT, WRECKAGE, CRASH, PLANET DESERT, ultra-detailed, film photography, light leaks, trending on artstation,\\ sharp focus, studio photo, intricate details, highly detailed, by greg rutkowski"}\\\toprule
Prompt 42 & \makecell[l]{``an expressionist charcoal sketch by Odilon Redon, drawing, face only, a gorgeous Japanese woman, hint of a smile,\\ noticeable charcoal marks, white background, no coloring, no color --ar 69:128 --s 750 --v 5. 2"}\\\toprule
Prompt 43 & \makecell[l]{``a man in a futuristic suit with neon lights on his face, cyberpunk art by Liam Wong, cgsociety, computer art, darksynth, synthwave, glowing neon"}\\\toprule
Prompt 44 & \makecell[l]{``The image features a bird perched on a branch, dressed in a suit and tie. The bird is holding a cup of hot coffee, in its beak. The coffee cup emits smoke.\\ The scene is quite unusual and whimsical. The bird's attire and the presence of the cup create a sense of humor and playfulness in the image."}\\\toprule
Prompt 45 & \makecell[l]{``carnage, a formidable supervillain, symbiote, bloody, psychopathic, unstoppable, mad, sharp teeth, epic composition, dramatic, gloomy, in the style\\ of mike deodato, realistic detail, realistic hyper-detailed rendering, realistic painted still lifes, insanely intricate"}\\\toprule
Prompt 46 & \makecell[l]{``A disoriented astronaut, lost in a galaxy of swirling colors, floating in zero gravity, grasping at memories, poignant loneliness, stunning realism, cosmic\\ chaos, emotional depth, 12K, hyperrealism, unforgettable, mixed media, celestial, dark, introspective"}\\\toprule
Prompt 47 & \makecell[l]{``an abstract painting of a beautiful girl, in the style of Pablo Picasso, masterpiece, highly imaginative, dada, salvador dali, i can't believe how beautiful\\ this is, intricate --ar 61:128 --s 750 --v 5. 2"}\\\toprule
Prompt 48 & \makecell[l]{``made by Emmanuel Lubezki, Daniel F Gerhartz, character of One Piece movie, Monkey D. Luffy, in straw hat, cinematic lighting, concept photoart,\\ 32k, photoshoot unbelievable half-length portrait, artificial lighting, hyper detailed, realistic, figurative painter with intricate details, divine proportion,\\ sharp focus, Mysterious"}\\\toprule
Prompt 49 & \makecell[l]{``a very detailed image of a female cyborg, half human, half machine, very detailed, with cables, wires, mechanical elements in the head and body,\\ dynamic light, glowing electronics, 4 k, inspired by H. r. Giger and Jean ansell and justin Gerard, photorealistic"}\\\toprule
Prompt 50 & \makecell[l]{``A beautiful photo of an lion that got lost in the amazon rainforest, rain, mist, 8k, sharp intricate details, masterpiece, imaginative, raytracing,\\ octane render, studio lighting, professionally shot nature photo, godrays, hyperrealistic, ultra high quality, realism, wet, dripping water,\\ wandering through the undergrowth"}\\
\bottomrule
\end{tabular}}
\end{table*}

\end{document}